\documentclass{article}

\usepackage{microtype}
\usepackage{graphicx}
\usepackage{subfigure}
\usepackage{booktabs} 

\usepackage{hyperref}

\usepackage[accepted]{icml2023}

\usepackage{amsmath}
\usepackage{amssymb}
\usepackage{mathtools}
\usepackage{amsthm}

\usepackage[capitalize,noabbrev]{cleveref}

\theoremstyle{plain}

\theoremstyle{definition}

\theoremstyle{remark}

\usepackage[textsize=tiny]{todonotes}

\icmltitlerunning{Simplified Temporal Consistency Reinforcement Learning}

\begin{document}

\twocolumn[\icmltitle{Simplified Temporal Consistency Reinforcement Learning}




\begin{icmlauthorlist}
\icmlauthor{Yi Zhao}{aalto-ee}
\icmlauthor{Wenshuai Zhao}{aalto-ee}
\icmlauthor{Rinu Boney}{aalto-cs}
\icmlauthor{Juho Kannala}{aalto-cs}
\icmlauthor{Joni Pajarinen}{aalto-ee}

\end{icmlauthorlist}

\icmlaffiliation{aalto-ee}{Department of Electrical Engineering and Automation, Aalto University, Finland}
\icmlaffiliation{aalto-cs}{Department of Computer Science, Aalto University, Finland}

\icmlcorrespondingauthor{Yi Zhao}{yi.zhao@aalto.fi}

\icmlkeywords{Machine Learning, Reinforcement Learning, Representation Learning, Planning, Model-based Reinforcement Learning, ICML}

\vskip 0.3in
]



\printAffiliationsAndNotice{}  

\begin{abstract}
Reinforcement learning (RL) is able to solve complex sequential decision-making tasks but is currently
limited by sample efficiency and required computation. To
improve sample efficiency, recent work focuses on model-based RL which interleaves model learning
with planning. Recent methods further utilize policy learning, value estimation, and, self-supervised learning as auxiliary objectives. 
In this paper we show that, surprisingly, a simple representation learning approach relying only on 
a latent dynamics model trained by latent temporal consistency is 
sufficient for high-performance RL. This applies when using pure planning with a dynamics model conditioned on the representation, but, also when utilizing the representation as policy and value function features in model-free RL.
In experiments, our approach learns an accurate dynamics model to solve challenging high-dimensional locomotion tasks with online planners while being 4.1$\times$ faster to train compared to ensemble-based methods. 
With model-free RL without planning, especially on high-dimensional tasks, such as the Deepmind Control Suite Humanoid and Dog tasks, our approach outperforms model-free methods by a large margin and matches model-based methods' sample efficiency while training 2.4{$\times$} faster.

\end{abstract}

\section{Introduction}
\label{intro}
\begin{figure}[t]
\centering
\includegraphics[width=0.41\textwidth, ]{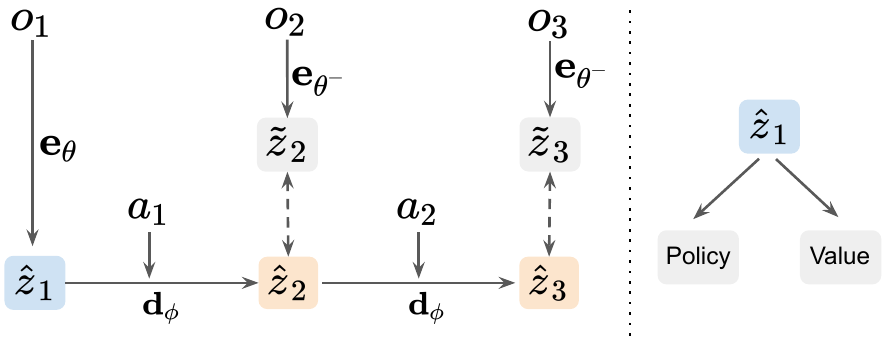}
\includegraphics[width=0.47\textwidth,]{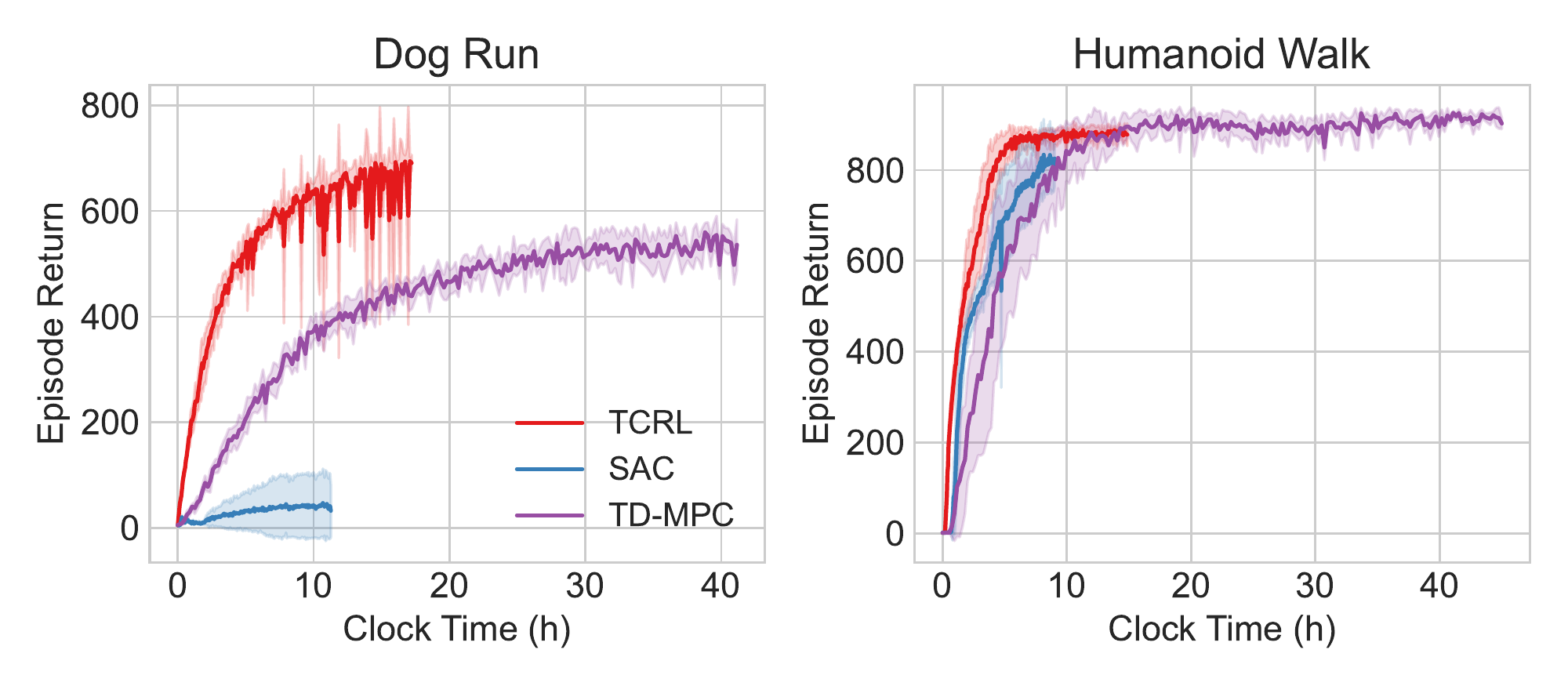}
\caption{\textbf{Top}: Model architecture for computing our temporal consistency reinforcement learning (TCRL) latent state $\hat{z}_t$. At each time step $t$ the model encodes observation $o_t$ into a latent state $\hat{z}_t$
using a neural network. The latent dynamics model $d_{\phi}$ predicts the next latent state $\hat{z}_{t+1}$ using $\hat{z}_t$ and action $a_t$. A standard momentum encoder $\textbf{e}_{{\theta}^-}$ prevents collapse of the latent state representation. We use the cosine loss between the latent state and the momentum encoded latent state, denoted by a dashed line, for training. The reward function, omitted for clarity, is trained with the standard MSE loss. This simple model works surprisingly well providing a trained dynamics model for planning experiments, and, in model-free RL experiments, the trained latent states $\hat{z}_t$ are used as inputs to policy and value functions.
\textbf{Bottom}: Episodic returns of TCRL compared to SAC and TD-MPC with respect to computing time on high-dimensional Humanoid and Dog tasks. Agents are trained with 5 million environment steps, and we plot 5 runs with shaded areas denoting 95\% confidence intervals.}

\label{fig: overview}
\end{figure}

Deep reinforcement learning (DRL) has shown promising results in games~\cite{schrittwieser2020mastering,DBLP:journals/jair/BellemareNVB13}, animation~\cite{peng2018deepmimic} and robotics~\cite{DBLP:conf/corl/WuEHAG22,DBLP:journals/ijrr/OpenAI20,lee2020learning}. However, DRL is data-demanding requiring millions of data points to train limiting the applicability of DRL in real-world scenarios.
To make DRL more sample-efficient, motivated by the success of self-supervised learning in both vision and language tasks, a series of recent works~\cite{DBLP:conf/corl/0006CHL20,DBLP:conf/iclr/SchwarzerAGHCB21,laskin2020curl} introduce self-supervised auxiliary losses. The aim is to improve representation learning for policy and value functions. The adopted self-supervised losses include image reconstruction \cite{DBLP:conf/aaai/Yarats0KAPF21,ha2018recurrent,watter2015embed} and maximizing the similarity between two augmentations of the same image (contrastive training)~\cite{laskin2020curl}. However, most of these methods demonstrate the effectiveness of the proposed method on pixel-based tasks. And their conclusions can not always directly transfer to state-based tasks, where a compact state representation is already available.

Recently, \citet{DBLP:conf/iclr/0001MCGL21,mcinroe2021learning,DBLP:conf/iclr/SchwarzerAGHCB21} exploit the temporal relationship of sequential observations and propose to facilitate representation learning by predicting future observations or future latent states. These methods usually train an image encoder and a latent dynamics model \emph{jointly} with the value function. 
\citet{DBLP:conf/icml/HansenSW22,schrittwieser2020mastering,ye2021mastering,DBLP:journals/corr/abs-2209-08466} extend these methods by leveraging the learned latent dynamics model to improve the policy and show promising results. For example, TD-MPC is the first documented method that solves high-dimensional dog tasks from DeepMind Control Suite~\cite{tunyasuvunakool2020} (DMC) by leveraging elaborate decision-time planning using the learned dynamics. 
Since all these methods learn and use the encoder, latent dynamics model, and value functions jointly, it is difficult to distinguish which parts contribute to performance improvements. 

This paper investigates the role of latent temporal consistency in state-based reinforcement learning. 
Specifically, we propose a simple but effective method, called TCRL (temporal consistency RL) to learn a state encoder and a latent dynamics model jointly, 
as shown in Figure \ref{fig: overview}. 
We carefully strip away complexities of previous methods and report empirical results on both i) dynamics model learning as well as ii) policy learning. 
We show that the trained model is accurate enough to be used for online planning to solve high-dimensional locomotion tasks. Compared to PETS~\cite{chua2018deep}, without using ensembles, probabilistic models, and state normalization, our method is able to solve high-dimensional DMC tasks, such as Quadruped Walk and Dog Walk ($\mathcal{A}\in \mathbb{R}^{38}, \mathcal{O}\in \mathbb{R}^{223}$) while being \textbf{4.1} $\times$ faster to train than ensemble-based methods which fail in these tasks. 
Also, even though in state-based tasks, inputs are already compact and have physical meaning, we still observe a big performance boost when training a \emph{model-free} agent in the learned latent space instead of the observation space. Our method improves the performance on challenging locomotion tasks compared to strong model-free baselines and matches sample efficiency compared to state-of-the-art model-based methods while being faster to train.
Furthermore, thanks to our minimalistic algorithm and implementation\footnote{Source Code of TCRL: \href{https://github.com/zhaoyi11/tcrl}{https://github.com/zhaoyi11/tcrl}}, we conduct extensive well-controlled experiments to analyze the usage of temporal consistency in state-based RL. We find that i) temporal consistency is more compatible with the cosine loss than the MSE loss; ii) joint training of the dynamics model and the value function introduces additional instability; iii) although our learned dynamics model is accurate to support online planning on high-dimensional tasks, we fail to observe strong benefits of using it in policy or value function learning.

\section{Related Work}
\paragraph{Learning dynamics models for planning} 
Learning an accurate dynamics model from interactions with the environment is crucial for model-based reinforcement learning (MBRL). When low-dimensional observations are accessible, it's common to learn a dynamics model in the observation space. PILCO~\cite{deisenroth2011pilco} and PETS~\cite{chua2018deep} stress the importance of capturing the model's uncertainty and explicitly incorporating the uncertainty into planning. Many recent MBRL methods~\cite{janner2019trust,DBLP:conf/iclr/ClaveraFA20,yu2020mopo,kidambi2020morel} employ the ensemble model used in PETS by default. However, its computational complexity grows linearly with the number of ensembles. And \citet{lutter2021learning} shows that these methods are struggling to handle high-dimensional tasks. 
Instead, we show that by leveraging the latent temporal consistency, we can learn a simple dynamics model to achieve better performance without using ensembles or probabilistic models.

\paragraph{Representation learning in RL} 
Representation learning has been investigated for decades in the context of RL. \citet{singh1994reinforcement,dearden1997abstraction,DBLP:conf/aaai/AndreR02,li2006towards,mannor2004dynamic,ruan2015representation,jiang2015abstraction,DBLP:conf/icml/AbelHL16} identify or learn a compact representation of state or action space but are usually limited to relatively simple environments. \citet{gelada2019deepmdp,DBLP:conf/iclr/0001MCGL21} extend these ideas to solve challenging tasks leveraging neural networks. Many recent works facilitate RL research by utilizing good insights from self-supervised learning (SSL). SSL aims to learn meaningful features without labels \cite{DBLP:journals/corr/KingmaW13,DBLP:conf/iclr/HigginsMPBGBML17,oord2018representation,DBLP:conf/icml/ChenK0H20,he2020momentum,grill2020bootstrap}. For example, \citet{hafner2019learning,DBLP:conf/iclr/HafnerLB020,DBLP:conf/iclr/HafnerL0B21,hafner2023mastering} learn good representations (latent space) in a VAE-like~\cite{DBLP:journals/corr/KingmaW13} way by reconstructing observations and then do planning and policy learning in the latent space. And \citet{DBLP:conf/icml/HansenSW22,DBLP:conf/iclr/SchwarzerAGHCB21,mcinroe2021learning} learn representations via contrastive learning~\cite{hadsell2006dimensionality}. SSL is mainly used in visual control tasks~\cite{DBLP:conf/aaai/Yarats0KAPF21,ha2018recurrent,laskin2020curl,huang2022accelerating}. However, the effectiveness of SSL in state-based tasks has not been clearly verified~\cite{yang2021representation}, since in this case, a compact representation with physical meaning is available. Also, \emph{none} of these methods try to investigate the accuracy of the learned dynamics model. Our paper aims to fill this gap.


\paragraph{Learning value-oriented dynamics models}
Recent work learns dynamics models without predicting future observations. \citet{silver2017predictron,oh2017value,tamar2016value,schrittwieser2020mastering,ye2021mastering} learn a dynamics model that only predicts rewards and values over multiple time steps and uses the learned model for planning. 
The state-of-the-art method TD-MPC~\cite{DBLP:conf/icml/HansenSW22}, closest work to ours, 
uses a temporal consistency constraint in the latent space.
TD-MPC shows good performance on continuous control tasks by training a latent dynamics model jointly with value functions and planning with the learned dynamics model.
Through extensive experiments, we argue \emph{none} of these are required to solve high-dimensional tasks such as the Humanoid and Dog tasks. 
Instead, we train the value function separately from the dynamics model and learn the policy and value function in a model-free way. Without using elaborate planning, our TCRL achieves competitive performance and is \textbf{2.4}$\times$ faster to train. Our results suggest that learning a high-quality state representation is a key factor in solving these challenging tasks.

\section{Method}
\paragraph{Components} As shown in Figure \ref{fig: overview}, our TCRL model includes four components:
\begin{equation}
\begin{aligned}
\text{Encoder} &: \ z_t = \mathbf{e}_{\theta}(o_t) \\
\text{Transition} &: \ z_{t+1}, r_t = \mathbf{d}_{\phi}(z_t, a_t) \\
\text{Value} &: \ q_{t} = \mathbf{q}_{\psi}(z_t, a_t) \\
\text{Policy} &: \ a_{t} \sim \pi_{\eta}(z_t)
\end{aligned}
\end{equation}
An encoder $\mathbf{e}_{\theta}$ maps an observation $o_t$ into a latent state $z_t$. In our method, two encoders are used, named online encoder $\mathbf{e}_{\theta}$ and momentum encoder $\mathbf{e}_{\theta^-}$. 
We calculate the target latent state $\tilde{z}_t$ using the 
momentum encoder with parameters $\theta^-$ which are the exponential moving average (EMA) of the online encoder parameters $\theta$. 
Based on the latent state $z_t$ and action $a_t$, a latent state at the next time step $z_{t+1}$ as well as an immediate reward $r_t$ are predicted. When inputs are pixels (see Appendix~\ref{appendix: pixels}), the encoder $\mathbf{e}_{\theta}$ is parameterized by a convolutional neural network (CNN), while MLPs are used when inputs are states. Also, both the latent dynamics model $\mathbf{d}_{\phi}$ and the value function $\mathbf{q}_{\psi}$ as well as the policy $\pi_{\eta}$ are represented by MLPs. 

In our method, the prediction happens in the \emph{latent} space instead of observation space. In section~\ref{result: plan}, we show that this is a key factor enabling us to learn an accurate dynamics model to support online planning. Furthermore, the value function and policy take the learned latent states instead of observations. In section~\ref{result: policy}, we suggest this is crucial for solving challenging high-dimensional Humanoid and Dog tasks. Although common in pixel-based tasks, our method explicitly demonstrates that in state-based tasks where a compact representation is already available, we can still benefit from using the latent states learned by temporal consistency.

\paragraph{Learning the encoder and latent dynamics}
The latent dynamics model is trained by accurately predicting $H$-step rewards $\tilde{r}_{t:t+H}$ and target latent states $\tilde{z}_{t+1:t+1+H}$. Specifically, for a $H$-step trajectory $(o_t, a_t, r_t, o_{t+1})_{t:t+H}$ drawn from the replay buffer $\mathcal{D}$, starting from an initial observation $o_t$, we first use the online encoder $\mathbf{e}_{\theta}$ to obtain a latent representation $\hat{z}_t = \mathbf{e}_{\theta}(o_t)$. Then, conditioning on action sequences $a_{t:t+H}$, the transition function $\hat{z}_{t+1}, \hat{r}_t =\mathbf{d}_{\phi}(\hat{z}_t, a_t)$ is applied iteratively to predict future rewards $\hat{r}_{t:t+H}$ and latent states $\hat{z}_{t:t+H}$.
The target rewards $\tilde{r}_{t:t+H}$ are just immediate rewards from the sampled trajectory, while the target latent states $\tilde{z}_{t:t+H}$ are calculated by the momentum encoder as $\tilde{z}_{t:t+H} = \mathbf{e}_{\theta^-}(o_{t:t+H})$.
Given multi-step predictions and targets, the training objective is simply to minimize the discounted sum of the MSE loss of rewards and the negative cosine distance of latent states:
\begin{equation}
    \sum_{h=0}^H \gamma^h \Big[||\hat{r}_{t+h} - \tilde{r}_{t+h}||_2^2 - \big(\frac{\hat{z}_{t+h}}{||\hat{z}_{t+h}||_2} \big)^{\top} \big(\frac{\tilde{z}_{t+h}}{||\tilde{z}_{t+h}||_2}\big) \Big].
    \label{eq: dynamics}
\end{equation}

This temporal consistency is used in several previous papers~\cite{DBLP:conf/iclr/SchwarzerAGHCB21,DBLP:conf/icml/HansenSW22,mcinroe2021learning}. Perhaps \citet{DBLP:conf/icml/HansenSW22} is the most related paper proposing the TD-MPC method. In addition to TD-MPC using planning at decision-time, we differ from TD-MPC in two aspects: i) we use the cosine loss but TD-MPC uses the MSE loss; ii) TD-MPC learns a latent dynamics model jointly with value functions. Learning value functions jointly with the dynamics model, so-called \emph{value-oriented} dynamics, enforces the dynamics model to encode strong task-specific information, which potentially makes it hard to generalize to new tasks~\cite{DBLP:conf/iclr/ZhangSP18}. Moreover, Section~\ref{ablation: objective} highlights additional training instability issues that joint training may cause.

\paragraph{Learning the policy and value function} 
For learning the policy and value function, we adopt a deep deterministic policy gradient method (DDPG)~\cite{DBLP:journals/corr/LillicrapHPHETS15} augmented with n-step returns~\cite{watkins1989learning,peng1994incremental} following~\citet{DBLP:conf/iclr/YaratsFLP22} with the only difference that instead of using the original observation $o_t$ we use the latent state $z_t = \mathbf{e}_{\theta}(o_t)$ as input to the policy and value function. Specifically, the value function is updated by minimizing:
\begin{equation} \label{eq: q}
\begin{aligned}
    \mathcal{L}_{\psi} &= \mathbb{E}_{\tau \sim \mathcal{D}} \big[ (\mathbf{q}_{\psi_k}(z_t, a_t) - y)^2 \big], \forall k \in {1,2} \\
    y &= \sum^{n-1}_{h=0}\gamma^h r_{t+h} + \gamma^n \min_{k=1,2} \mathbf{q}_{\psi^-_k}(z_{t+n}, a_{t+n})
\end{aligned}
\end{equation}
where $a_{t_n} = \pi_\eta (z_{t+n}) + \epsilon$ and the noise $\epsilon$ is sampled from a clipped Gaussian distribution $\epsilon \sim \mathcal{N}(0, \sigma^2)$ with a linearly decayed $\sigma$ as in DrQv2\cite{DBLP:conf/iclr/YaratsFLP22}. Following previous model-free methods \cite{fujimoto2018addressing,haarnoja2018soft,hasselt2010double}, we use double Q networks $\mathbf{q}_{\psi_{1,2}}$ as well as two delayed Q networks $\mathbf{q}_{\psi^-_{1,2}}$. 
We update the actor's parameters using the loss
\begin{equation} \label{eq: policy}
    \mathcal{L_{\eta}} = -\mathbb{E}_{\tau \sim \mathcal{D}}\big[ \min_{k=1,2} \mathbf{q}_{\psi_k}(z_t, a_t) \big]
\end{equation}
that maximizes Q-value for the actor.
Again, $z_t = \mathbf{e}_\theta (o_t)$, action $a_t = \pi_\eta(z_t) + \epsilon$ and we do not update the encoder's parameters with actor's gradients.

\section{Experiments}
\begin{figure*}[t]
\centering
\includegraphics[width=0.95\textwidth]{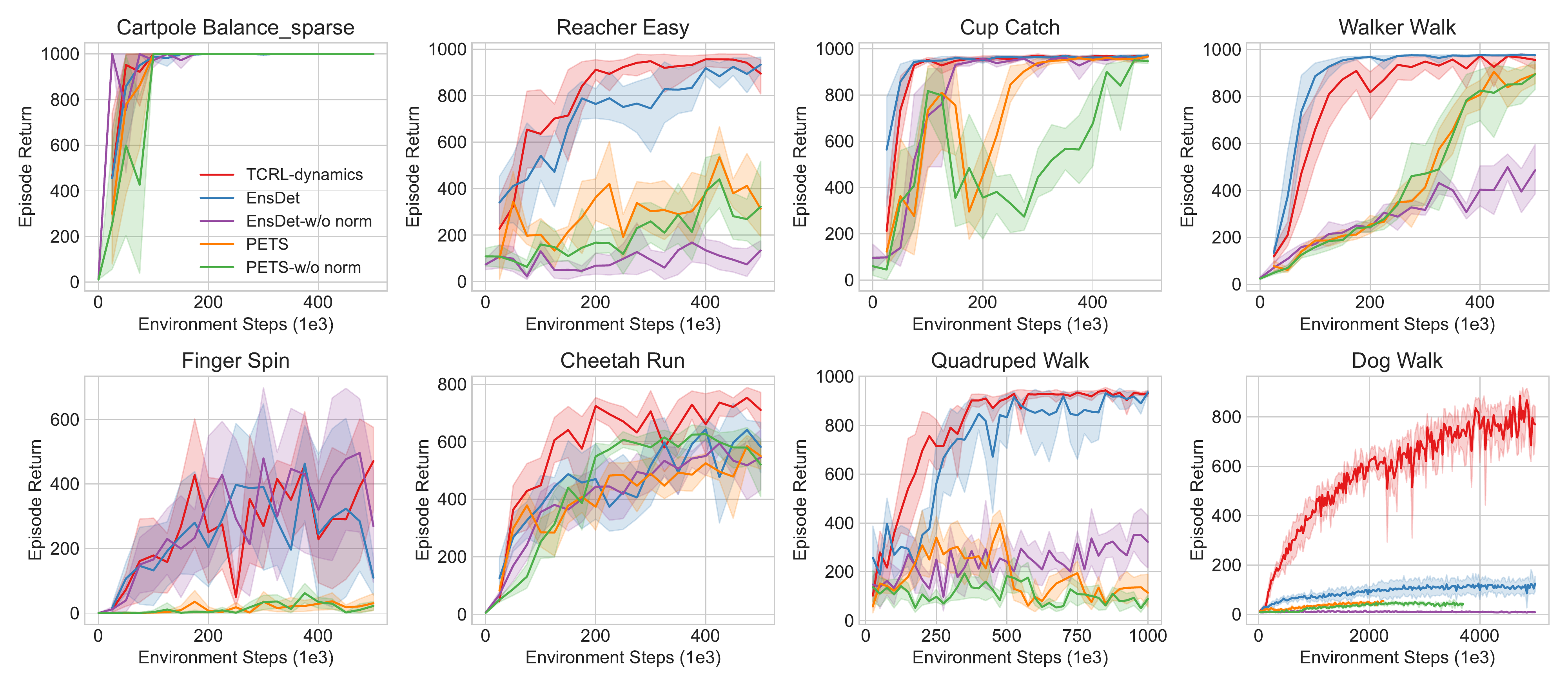}
\caption{
Planning performance on DMC continuous control benchmarks. We plot the mean across 5 runs. Each run includes 10 trajectories. Shaded areas denote 95\% confidence intervals. Without using ensembles, our TCRL-dynamics consistently outperforms ensemble-based methods w.r.t.~planning performance but with $\sim 4\times$ less training time. Especially on the challenging Dog Walk task, TCRL-dynamics outperforms previous methods by a large margin, which makes it a good candidate to learn the dynamics model in model-based RL.}
\label{fig: result_plan}
\end{figure*}

We evaluate our TCRL approach in several continuous DMC control tasks. DMC uses scaled rewards, where the maximal episodic returns are 1000 for all tasks. 
We evaluate our method in \emph{two} separate settings:
\begin{itemize}
    \setlength\itemsep{0em}
    \item \textbf{Dynamics learning.} In the first setting, there is \emph{no} value function or policy involved. Instead, we directly use the latent dynamics model for online planning calling the approach \emph{TCRL-dynamics}. The aim is to answer whether temporal consistency can be used to learn an accurate latent dynamics model.
    \item \textbf{Policy learning.} In the second setting, we train the policy and value functions in a model-free way using latent states in place of original observations calling the approach \emph{TCRL}. This experiment aims to investigate whether the latent representation trained by temporal consistency benefits policy and value function learning. 
\end{itemize}

\subsection{Evaluation of the Latent Dynamics Model} 
\paragraph{Evaluation metrics}
Learning an accurate dynamics model is critical for model-based RL research. Finding a proper way to evaluate the learned dynamics model is still an open problem. Unlike usual supervised learning, the mean square error on the test set does not directly reflect the model's performance in planning~\cite{lutter2021learning,DBLP:conf/l4dc/LambertAYC20}.

Considering the primary usage of a dynamics model is planning, we directly evaluate the learned model with its planning results using Model Predictive Path Integral (MPPI)~\cite{camacho2013model,williams2015model}. 
MPPI is a population-based \emph{online} planning method that iteratively improves the policy $a_{t:t+H}$ with samples. In each iteration $j$, $N$ trajectories are sampled according to the current policy $a^j_{t:t+H}$. Then, $K$ trajectories with higher returns $\sum_{h=0}^H r_{t+h}^j(s^j_{t+h}, a^j_{t+h})$ are selected. Next, we calculate an improved policy $a^{j+1}_{t:t+H}$ by taking the weighted average over the selected top-K trajectories, where the weights are calculated by taking the softmax over returns of the top-K trajectories. After $J$ iterations, the first action of $a^{J}_{t:t+H}$ is executed. 

In our MPPI implementation, \emph{no} value function or policy network is involved.
Instead, we iteratively learn the dynamics model by i) collecting experiences via the MPPI planner, and ii) improving the dynamics model using collected data by optimizing Equation~\ref{eq: dynamics}.
In this way, all sampled action sequences are evaluated by the learned latent dynamics model alone, thus the planning performance can reflect the model's accuracy. 
We consider the following comparison methods to learn the dynamics model:
\begin{itemize}
    \setlength\itemsep{0em}
    \item \textbf{Stochastic ensemble model} (PETS) learns an ensemble of stochastic neural networks predicting both mean and standard deviation of next states and rewards. Similarly to \citet{chua2018deep}, we only predict the one-step future since uncertainty propagation through multiple time steps is unclear~\cite{lutter2021learning}. PETS is commonly used in many model-based RL methods when inputs are in compact states. 
    \item \textbf{Deterministic ensemble model} (EnsDet) uses an ensemble of deterministic neural networks to predict \emph{multi-step} future observations and rewards. EnsDet's architecture is similar to our method with the difference of predicting the next observations instead of the next latent states enabling an experimental comparison between observation predictions and latent space predictions.
\end{itemize}
Although our primary focus is on state-based tasks, we also test our method on pixel-based tasks to show that it is general. In the experimental comparison with PlaNet~\cite{hafner2019learning} in Appendix~\ref{appendix: pixels}, our method obtains comparable or better performance on six commonly used pixel-based benchmark tasks. Detailed ablation studies on pixel-based tasks are in Appendix~\ref{appendix: pixel, ablation}.

\begin{figure*}[t]
\centering
\includegraphics[width=0.95\textwidth]{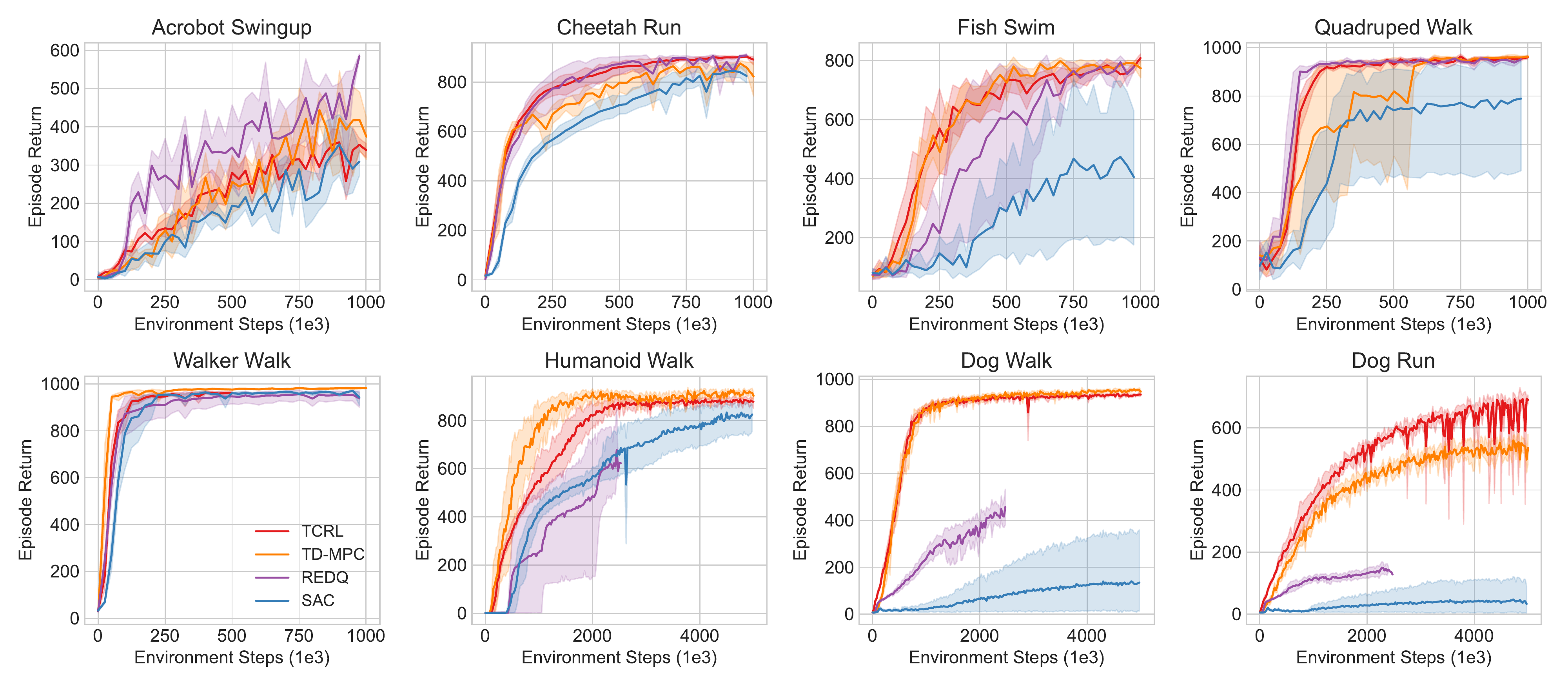}
\caption{Policy performance on eight selected DMC tasks. We plot the mean across 5 runs, and each run includes 10 trajectories. The shaded areas denote 95\% confidence intervals. Our TCRL outperforms strong model-free baselines SAC and REDQ by a large margin on challenging Humanoid and Dog locomotion control tasks. We also achieve comparable performance compared to the state-of-the-art model-based method TD-MPC without using elaborate planning. REDQ and TD-MPC take $18.3 \times$ and $2.4 \times$ training time respectively compared to TCRL. Full results on 24 continuous control tasks are shown in Appendix~\ref{result: full}.}
\label{fig: result_policy}
\end{figure*}

\paragraph{Planning results} 
\label{result: plan}
In Figure \ref{fig: result_plan}, we compare our method with the stochastic ensemble model (PETS) and deterministic ensemble model (EnsDet) on eight state-based tasks from DMC, including six commonly used benchmarking tasks and two challenging high-dimensional tasks: Quadruped Walk and Dog Walk. Although simple and without using ensembles, our method either matches or outperforms comparison methods in the tested tasks.
The high-dimensional action and observation spaces $\mathcal{A}\in \mathbb{R}^{38}, \mathcal{O}\in \mathbb{R}^{223}$ make solving the Dog Walk task, even for a strong model-free baseline~\cite{haarnoja2018soft}, difficult. In fact, to the best of our knowledge, TCRL-dynamics is the first documented method that can control the Pharaoh Dog walking forward using online planning with the learned dynamics model. 

\citet{chua2018deep} discuss the importance of predicting both aleatoric and epistemic uncertainty to achieve good planning performance. However, in both \citet{chua2018deep} and our experiments, the algorithms are tested in deterministic environments, which makes predicting the aleatoric uncertainty less motivated. 
In fact, in our experiments, TCRL-dynamics and EnsDet outperform PETS on all tasks. 
This evidence forces us to rethink the role of predicting aleatoric uncertainty in these commonly used deterministic tasks.
Also, compared to PETS, both our method and EnsDet show the importance of using the multi-step training objective while it is not well compatible with the stochastic model used in PETS since properly propagating the uncertainty over multi-steps is challenging. Compared to EnsDet, our results show the superiority of predicting in the latent space instead of the observation space. We discuss this more in Section~\ref{ablation: latent}. 

Moreover, we find that both PETS and EnsDet require state normalization. However, state normalization is not common in off-policy methods as this may introduce additional instability. Our assumption is that for off-policy methods, the data distribution in the replay buffer keeps changing during training which leads to the unstable mean and standard deviation for normalizing states. But TCRL-dynamics achieves good performance without state normalization, making it attractive to be adopted in model-based RL.

As mentioned before, we also evaluate our method on pixel-based tasks (Appendix~\ref{appendix: pixels}). On six visual control tasks, we show that our simple method matches PlaNet's results, which require an RNN and separated deterministic and stochastic paths to model the dynamics. Also, our method is easy to implement and \textbf{5.51} $\times$ faster to train. Our results show that this simple model can be competitive to be used in both pixel-based and state-based MBRL. 

\subsection{Evaluation of the Policy} \label{result: policy}
We show that predicting in the latent space enables us to learn an accurate dynamics model to support online planning. Now, we try to introduce a policy and value to solve challenging tasks. We evaluate our TCRL methods on 24 continuous control tasks from DMC. Please check Appendix \ref{result: full} for full results. We select 8 representative tasks to compare with other both model-based and model-free methods w.r.t sample efficiency and compute efficiency.
We consider the following methods in our experiments:
\begin{itemize}
    \setlength\itemsep{0em}
    \item \textbf{TD-MPC} achieves promising performance on challenging tasks with \emph{value-oriented} latent dynamics model, where the latent dynamics model is learned jointly with the value function, as well as elaborate decision-time planning. We include TD-MPC to demonstrate that $none$ of these techniques are crucial to achieving good performance on challenging tasks, but state representation matters. 
    \item \textbf{SAC} is a strong model-free baseline that achieves good performance among different DMC tasks. We include SAC to show how much the policy learning benefits from the state representation even when compact state representation is available.
    \item \textbf{REDQ}~\cite{DBLP:conf/iclr/ChenWZR21} is a strong model-free baseline that is built upon SAC but with a higher Update-To-Data ratio. We include it because this model-free method outperforms many strong model-based methods, such as PETS~\cite{chua2018deep}, MBPO~\cite{janner2019trust}, and MVE~\cite{feinberg2018model}, with respect to sample efficiency.
\end{itemize}
We also compared ALM~\cite{DBLP:journals/corr/abs-2209-08466}, a recent model-based approach learning representation, latent-space model, and policy jointly. But we fail to achieve good performance in DMC benchmark environments possibly due to hyperparameters in the official ALM implementation being tuned for OpenAI Gym~\cite{brockman2016openai}. We include ALM results in Appendix~\ref{appendix: baseline}.

\paragraph{Policy results} 
Figure~\ref{fig: result_policy} compares the performance of TCRL with strong baselines on eight continuous control tasks, including challenging Humanoid and Dog domains. TCRL outperforms SAC on all tested tasks. Especially on complex tasks, such as Fish Swim, Humanoid Walk, Dog Walk, and Dog Run, our method outperforms SAC by a large margin. Since we also learn policy and value functions in a model-free way similar to SAC, the major performance improvement is from the state representation. Our results show strong evidence that state representation is critical even when a compact state representation is available, and temporal consistency can extract useful features that benefit policy and value function learning. 

REDQ uses randomized ensembled double Q-learning to achieve a high update-to-data ratio. It achieves better performance than SAC on most tasks except the Humanoid Walk task. TCRL outperforms or matches REDQ on all tasks except the Acrobot Swingup task without increasing the update-to-data ratio and thus being $18.3 \times$ faster to train. Notice that TCRL and REDQ improve sample efficiency via orthogonal ways and can be easily combined.

To recapitulate, compared to TD-MPC, TCRL i) uses a cosine loss on latent states; ii) achieves comparable performance without a value-oriented latent dynamics model; iii) does not perform decision-time planning. The results show that using temporal consistency to extract useful state representations is a key factor to success in solving Humanoid and Dog tasks. Because we do not use decision-time planning, which is computationally heavy, our method is $2.4 \times$ faster to train. Furthermore, unlike TD-MPC, our method does not need to select pre-task action repeat (control frequency), making it easier to use.
Moreover, Section~\ref{ablation: objective} analyzes why TCRL achieves better performance on the Dog Run task.

\section{Empirical Analysis} \label{ablation study}
This section empirically studies TCRL with special attention to answering: i) why does TCRL work well? ii) which training objective should be used? iii) can we use the learned dynamics in policy learning?
More comparison studies regarding hyperparameters and the choice of different training objectives are shown in Appendix~\ref{appendix: ablation}.

\paragraph{Why does TCRL work well?} \label{ablation: latent}
\begin{figure}[t]
\centering
\includegraphics[width=0.48\textwidth]{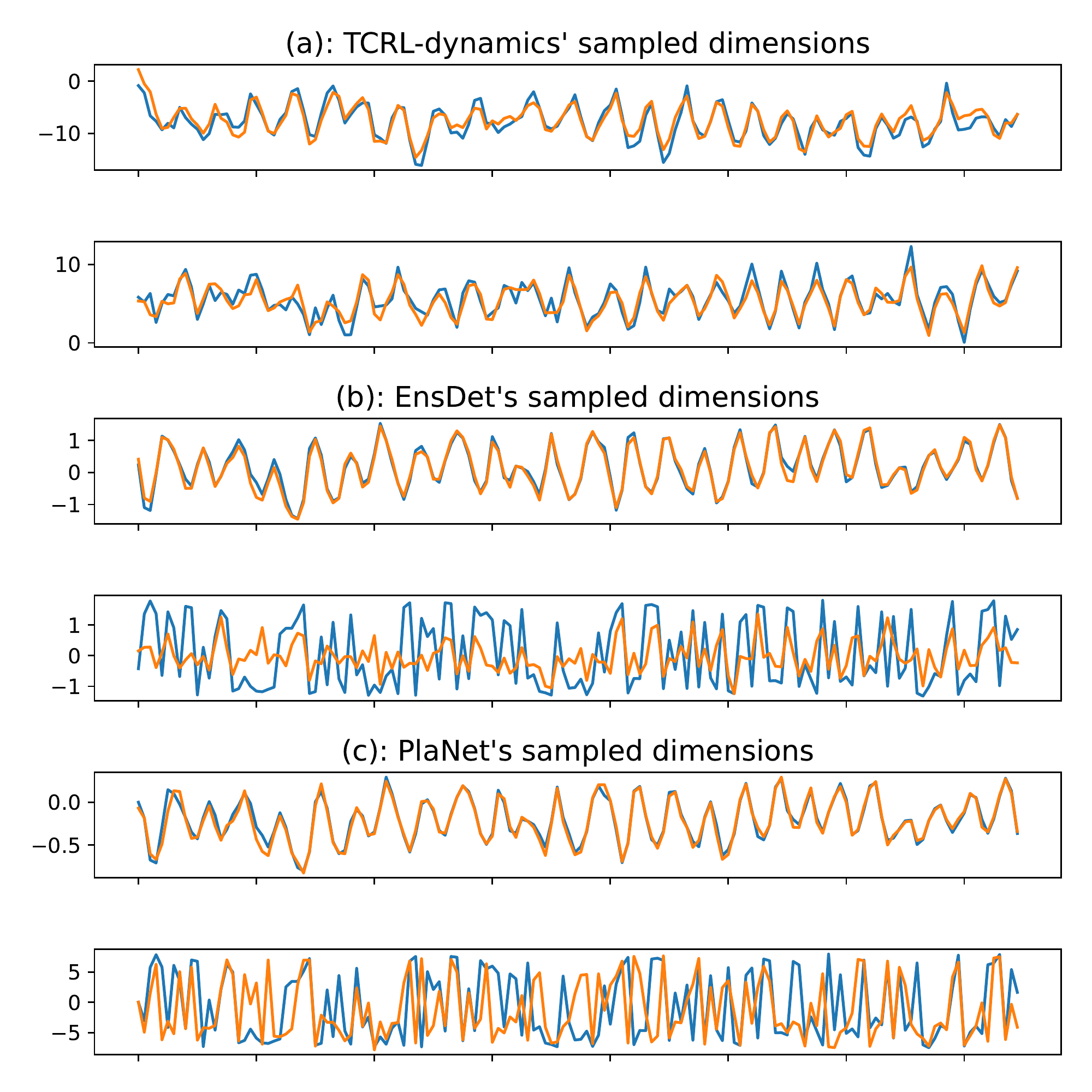}
\caption{Comparison of open-loop predictions. Blue curves are ground truths and orange curves are predicted by learned dynamics models. \textbf{Top:} two sampled dimensions $z_t$ from a Quadruped Walk trajectory in the latent space learned by TCRL-dynamics. \textbf{Middle:} two sampled dimensions $o_t$ from the same trajectory but in the normalized observation space. The orange curves are predicted by EnsDet. \textbf{Bottom:} two sampled dimensions $o_t$ from the same trajectory and the orange curves are reconstructed observations from the PlaNet model.}
\label{fig: state_repr}
\end{figure}

Experiments in dynamics model learning and policy learning show that one driving factor of TCRL's success is temporal consistency training extracting a useful state representation. 
Although the state from the simulator is already a compact representation and has physical meanings, the state still suffers from several possible drawbacks: i) not translation invariant; ii) not rotation invariant and the double-cover issue of quaternions; iii) magnitude differences between different dimensions; iv) strong temporal discontinuity.

The first two drawbacks are partially avoided by handcrafted features, e.g., DMC removes the Cheetah's position from the observation and represents angles by their sine and cosine in the Cartpole. For the third one, perhaps the most straightforward solution is to normalize the states. As shown in Figure~\ref{fig: result_plan}, normalizing states introduces an obvious planning performance boost. However, as mentioned before, normalizing the states may introduce additional instability, and most off-policy methods~\cite{haarnoja2018soft,fujimoto2018addressing,DBLP:journals/corr/LillicrapHPHETS15} do not utilize normalization. The fourth drawback is caused by the nature of locomotion tasks, that is when a robot contacts the ground, the acceleration and sensor readings change significantly.

In Figure~\ref{fig: state_repr}, we compare open-loop predictions of the TCRL-dynamics, EnsDet, and PlaNet. Given an initial observation and an action sequence, we predict the future states up to 150 steps with different methods. The blue curves are ground truths while the orange curves are predictions. All plots are from the same Quadruped Walk trajectory but with different sampled dimensions. Figure~\ref{fig: state_repr}(a) shows the open-loop predictions in the latent space learned by TCRL-dynamics. Figure~\ref{fig: state_repr}(b) plots the predictions in the normalized observation space by EnsDet. Figure~\ref{fig: state_repr}(c) shows reconstructed observations from PlaNet. Qualitatively our learned latent space is smoother and future latent states are easier to predict. The magnitude differences between dimensions and temporal discontinuity make it hard to train an accurate dynamics model by EnsDet and PlaNet.

Also, our method can ease optimization, according to~\citet{tian2021understanding}, the exponential moving average of the momentum encoder can be seen as an automatic curriculum. In the beginning, the training target $\tilde{z}_t$ is close to the prediction setting an easier target for training. Then, the target gradually becomes hard and then tends to stabilize as it converges. 

\paragraph{Training objective} \label{ablation: objective}
\begin{figure}[t]
\centering
\includegraphics[width=0.48\textwidth]{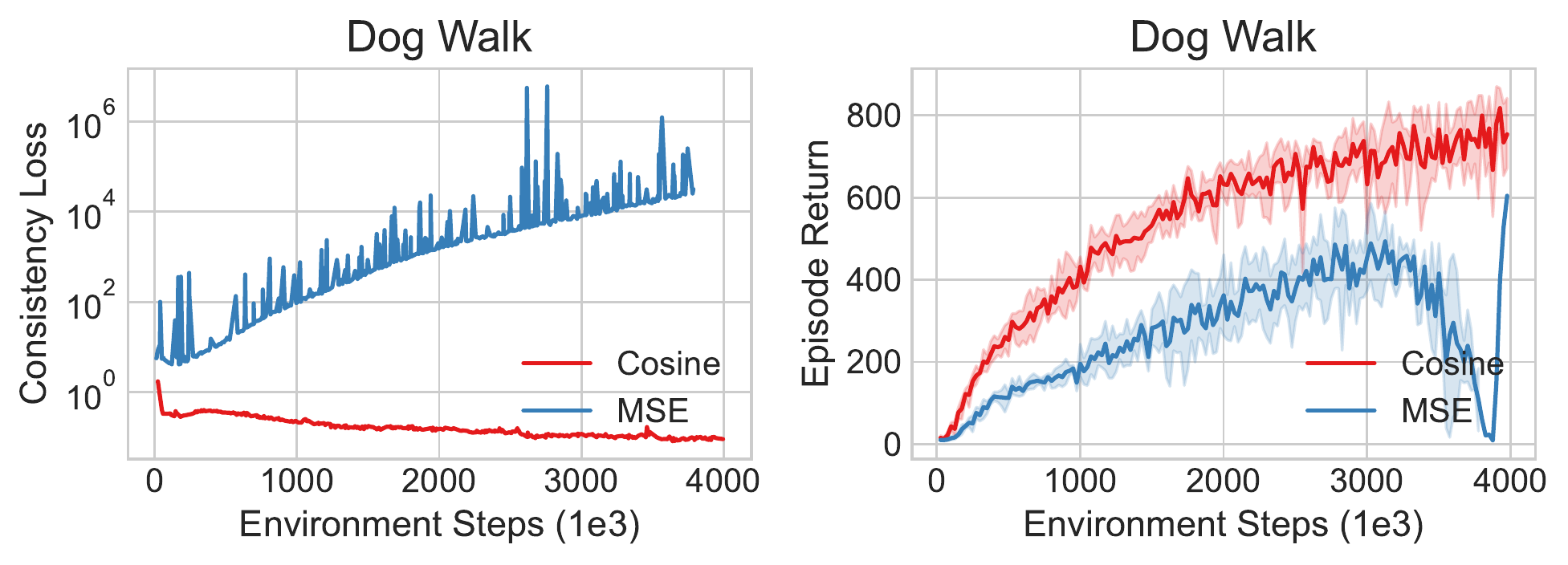}
\includegraphics[width=0.48\textwidth, ]{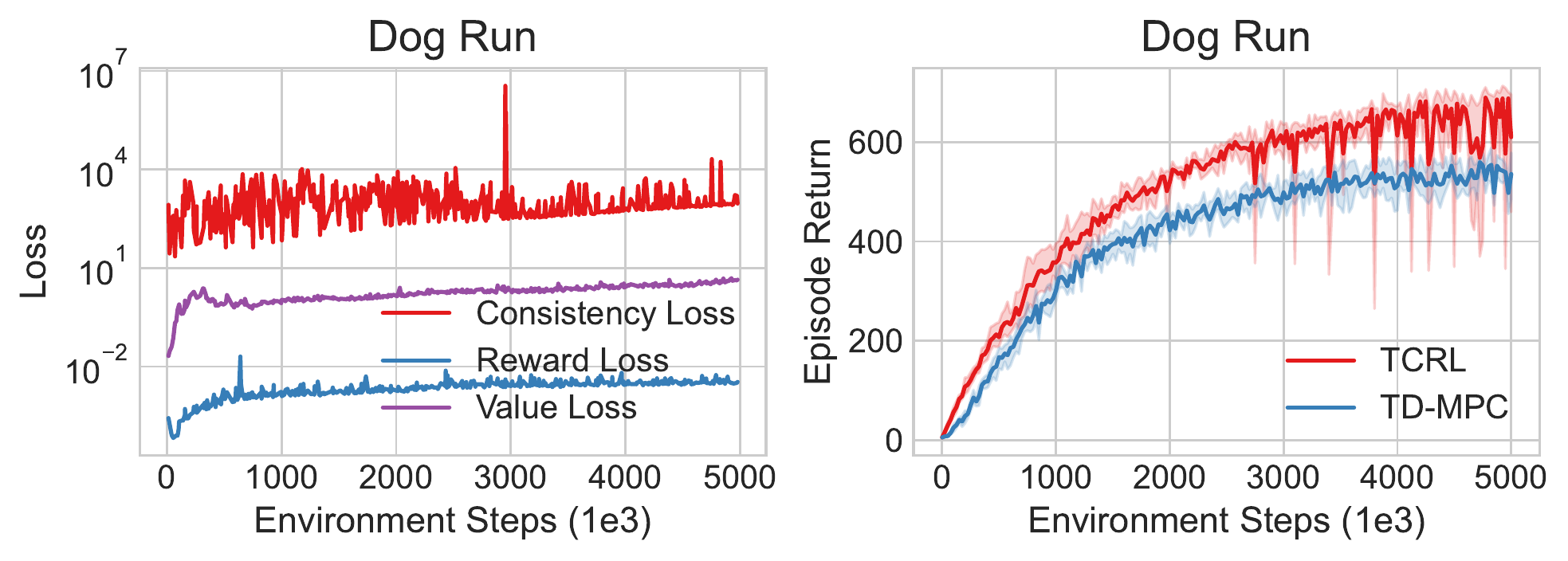}
\includegraphics[width=0.48\textwidth]{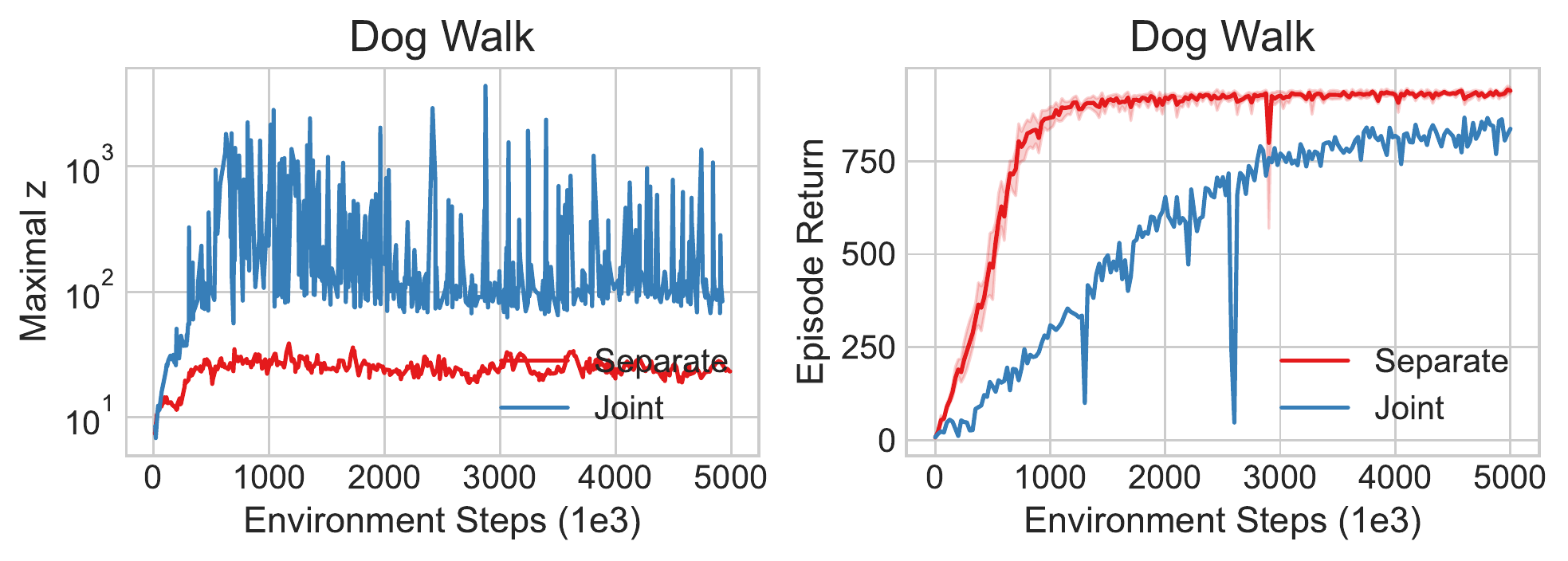}
\caption{Comparison of choosing different training objectives. \textbf{Top}: we compare MSE with cosine loss in Equation~\ref{eq: dynamics} in our \emph{TCRL-dynamics} variant. \textbf{Middle}: when jointly training the value function and the dynamics model with the MSE loss, as in TD-MPC, the consistency loss dominates the loss and leads to worse performance. \textbf{Bottom}: when jointly training the value function and the dynamics model with the cosine loss, the latent space keeps expanding, leading to performance degeneration.}
\label{fig: ablation, objective}
\end{figure}
\begin{figure*}[t]
\centering
\includegraphics[width=0.95\textwidth, ]{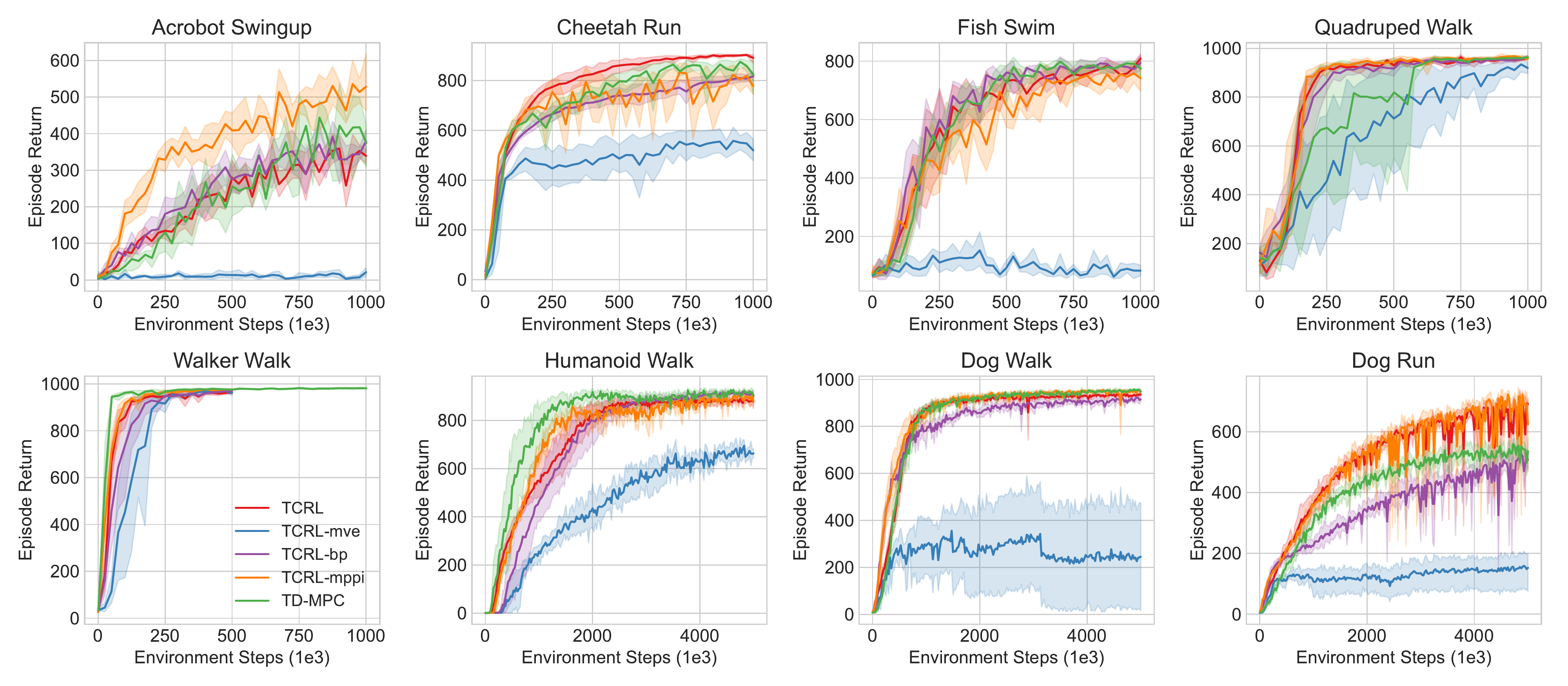} 

\caption{Comparison of different ways to use the learned dynamics in policy or value function learning.}
\label{fig: ablation, dynamics}
\end{figure*}

As mentioned before, different from TD-MPC, we use cosine loss to measure the distance between the predicted and the target latent states and train a latent dynamics model separately with value functions. We now analyze our design choices.

For the choice of the loss function, previous self-supervised learning methods commonly use the cosine loss~\cite{he2020momentum,grill2020bootstrap} and demonstrate performance drops with the MSE loss. The same evidence appears in RL setting~\cite{DBLP:conf/iclr/SchwarzerAGHCB21}. We found that in the \emph{TCRL-dynamics} variant, where \emph{no} policy or value function is involved, using MSE loss works well in most tasks. However, we do observe a severe performance decrease on the Dog Walk task where observations are complex as in the top row of Figure~\ref{fig: ablation, objective}. With the MSE loss, training tends to be unstable instead of collapse~\cite{DBLP:conf/iclr/SchwarzerAGHCB21}. Whilst TCRL-dynamics, trained with cosine loss, is much more stable during training and achieves better asymptotic performance.

We now discuss the case when jointly training value functions and a dynamics model. When the MSE loss is used as in TD-MPC, we found that training is stabilized due to the regularization from value function learning. However, as shown in the middle row of Figure~\ref{fig: ablation, objective}, the consistency loss (shown in the log scale) will dominate the overall loss. This will lead to a sub-optimal solution as observed in the Dog Run task as well as visual control tasks~\cite{DBLP:conf/icml/HansenSW22}.

Since we found that cosine loss works better than MSE loss in the TCRL-dynamics variant, a natural idea is to use cosine loss to replace the MSE loss in TD-MPC, which can potentially solve the dominating issue. However, as shown in the bottom right plot of Figure~\ref{fig: ablation, objective}, a worse performance is observed. To explain it, we should notice that learning a value function is not supervised training. Many ways have been proposed to stabilize training, but perturbation during training still exists. 
The bottom left plot of Figure~\ref{fig: ablation, objective} shows the maximal latent state value (in the log scale) in training batches. As we can see, the magnitude of the latent space keeps growing. When training the value function and the latent dynamics model separately as in TCRL, the latent space is much more stable and achieves better performance.

Except for training stability, training a value function and dynamics model separately potentially enables better generalization ability as the encoder does not encode strong task-specific features from the value function~\cite{sekar2020planning,DBLP:conf/iclr/ZhangSP18}. We will leave this as future work.

\paragraph{Utilizing the dynamics model in policy training} 
Here, we use the learned latent dynamics model to train the policy and value function by
i) model-based value expansion (TCRL-mve)~\cite{feinberg2018model}, where we use the dynamics model to rollout a short trajectory and calculate k-step returns used to train the value function; ii) updating the policy by backpropagating through the dynamics model (TCRL-bp)~\cite{deisenroth2011pilco,DBLP:conf/iclr/HafnerLB020}; iii) policy-guided decision-time planning as in TD-MPC (TCRL-mppi). 

Although the TCRL-dynamics variant showed that our learned latent dynamics model is accurate to support planning, Figure~\ref{fig: ablation, dynamics} shows performance drops with the TCRL-mve and TCRL-bp variants, which is a common issue in model-based RL~\cite{feinberg2018model,chua2018deep}. In RL, the policy tends to exploit model errors decreasing asymptotic performance. We found that on most tasks, TCRL-bp achieves reasonable performance, but performance drops happen on the Dog Run task. However, as shown in Figure~\ref{fig: ablation, dynamics}, decision-time planning (TCLR-mppi) benefits from the learned dynamics model achieving better performance on the Acrobot Swingup and Dog Run tasks than both TCRL and TD-MPC. TCLR-mppi is a safe way to utilize the learned dynamics model as the dynamics model only influences the collection of informative samples, but is not directly involved in policy or value function updating. Note that TCLR-mppi does not use prioritized experience replay~\cite{DBLP:journals/corr/SchaulQAS15} and pre-task action repeat (control frequency) as used in TD-MPC, which may in some tasks lead to slightly worse performance than TD-MPC, such as in Cheetah Run. 

\section{Conclusion}
In this paper, we propose a simple yet efficient way to learn a latent representation based on temporal consistency. We show that the learned representation benefits policy learning to solve challenging state-based Humanoid and Dog tasks, outperforming model-free methods by a large margin and matching model-based methods with respect to sample efficiency but being 2.4 $\times$ faster to train. Furthermore, we show that the learned latent dynamics model is accurate enough to support online planners to solve high-dimensional tasks, outperforming previous ensemble-based methods but being much faster to train. Yet we believe that our method can be improved in different ways, for example, by using the learned model to improve the policy or value function learning, or extending to visual control tasks.

\section*{Acknowledgements}

We acknowledge the computational resources provided by the Aalto Science-IT project and CSC, Finnish IT Center for Science, and, funding by Academy of Finland (345521).

\bibliography{example_paper}
\bibliographystyle{icml2023}

\newpage
\appendix
\onecolumn

\section{Planning Performance in Pixel-based Tasks} \label{appendix: pixels}

\subsection{Pixel-based Control Tasks}
\begin{figure*}[th]
\centering
\includegraphics[width=0.85\textwidth, ]{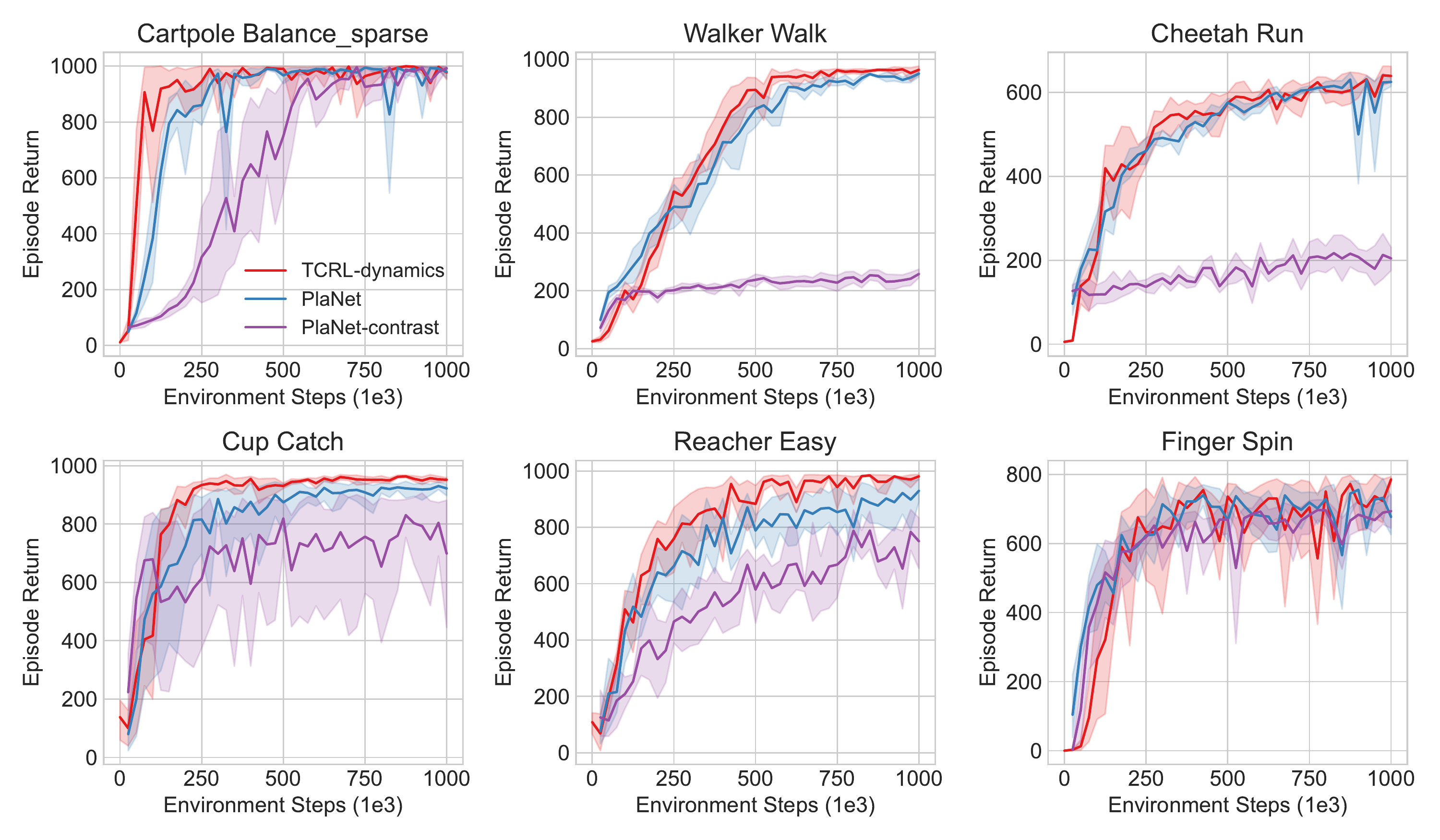}
\caption{Planning performance on pixel-based DMC benchmarks. We plot the mean across 5 runs. Each run includes 10 trajectories. Shaded areas denote 95\% confidence intervals. Our TCRL-dynamics matches PlaNet while being 5.51{$\times$} faster to train. TCRL-dynamics outperforms PlaNet with contrastive objective by a large margin.}
\label{fig: pixel}
\end{figure*}

In Figure \ref{fig: pixel}, we compare TCRL-dynamics with PlaNet~\cite{hafner2019learning} and PlaNet with contrastive loss~\cite{hafner2019learning,DBLP:conf/corl/0006CHL20}. To extend TCRL-dynamics from state-based tasks to visual control tasks, we use a convolutional neural networks (CNN) to represent the encoder and inputs are three stacking images. Following \citet{DBLP:conf/aaai/Yarats0KAPF21}, we use random crop data augmentation.  
PlaNet is a strong baseline in pixel-based tasks. It uses the recurrent state space model (RSSM), which uses an RNN and separated deterministic and stochastic paths to model the dynamics. PlaNet-contrast replaces the sequential VAE-like loss in PlaNet with a contrastive loss.
We re-implement PlaNet using Pytorch~\cite{paszke2019pytorch} to have a fair computational time comparison. To match our method, the learning rate is changed from 1e-3 to 5e-4, and the state dimension is increased from 30 to 50. Furthermore, we use MPPI instead of CEM as the planner to select actions. We use the same hyperparameters as in state-based tasks listed in Appendix \ref{appendix: hyperparameters}, but increase the population, elite size and iteration to 1000, 100 and 10 respectively. We verify that, with MPPI, our implementation outperforms the original results~\cite{hafner2019learning}. 

Though our method only uses tacking images as inputs and adopts a simple architecture, it achieves competitive performance on most tasks and outperforms PlaNet on Cup Catch and Reacher Easy tasks. 
In these two tasks, the ball or the goal position only takes tens of pixels. As mentioned by~\citet{okada2021dreaming}, the RSSM can still obtain low reconstruction error even totally ignoring these pixels. Our method does not suffer from this issue since we do not reconstruct the observations. Our method also obviously outperforms PlaNet-contrast, showing the effectiveness of the temporal consistency objective. 

Furthermore, due to the simplicity of our method, it is much faster to train. We evaluate the training time on a single RTX 2080Ti GPU. On the Cartpole Blance\_sparse task, PlaNet takes 15.6 hours for 500 episodes while TCRL-dynamics only takes \textbf{2.83} hours, which is \textbf{5.51} $\times$ faster than PlaNet. We hope our results can speed up model-based RL research.

\subsection{Distracting Control Tasks}\label{appendix: pixel,distracting}

In the real world, observations are complex, while observations from DMC have plain backgrounds making it easy to distinguish interested objects. To test the performance of the learned model under complex observation, we test our methods as well as two pixel-based baselines using distracting control suite~\cite{stone2021distracting}. Distracting control suite modifies upon standard DMC by adding natural backgrounds, color, and camera distractions. In our experiments, only natural background distractions are considered. Images of background distractions are video frames sampled from the DAVIS dataset \cite{pont20172017}. In each episode, the background will be re-sampled from the dataset. We consider a simpler case where we only sample images from four pre-sampled videos.

\begin{figure*}[t]
\centering
\includegraphics[width=0.85\textwidth, ]{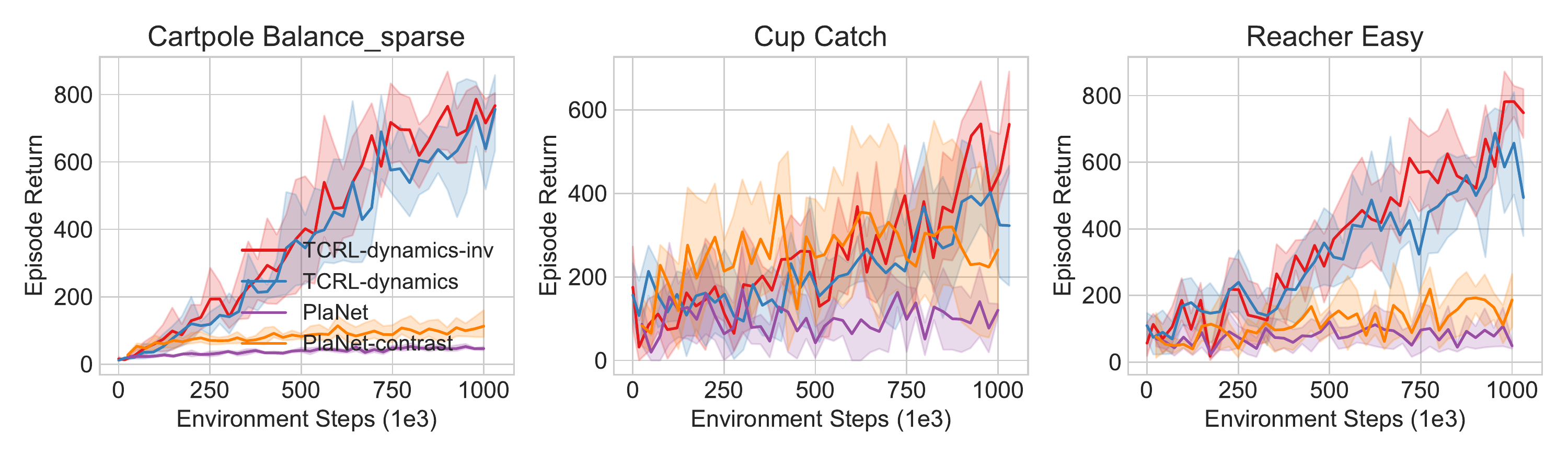}
\caption{Planning results on distracting pixels-based tasks. We plot the mean across 5 runs and the shaded areas denote 95\% confidence intervals. TCRL-dynamics outperforms PlaNet and PlaNet-contrast by a large margin.}
\label{fig: distracting}
\end{figure*}

To enable the dynamic model to focus on controllable objects, we add an additional \emph{inverse dynamic model}~\cite{pathak2017curiosity}. The inverse dynamic model takes two adjacent latent states $z_t, z_{t+1}$ as inputs and predicts the corresponding action $a_t$.  
As shown in Figure ~\ref{fig: distracting}, our method is more robust to background distraction compared to PlaNet and PlaNet-contrast. We notice that PlaNet fails to solve these tasks completely. This is because PlaNet learns the dynamic model based on a VAE-like loss, which means it reconstructs the visual observation. However, on this distracting control suite, the backgrounds keep changing, making it hard to model. Also, only a few pixels are related to the control task, which can be easily ignored by PlaNet. While our method does not require to reconstruct observations, making it more robust in the distracting tasks. Furthermore, we observe that the inverse dynamic model helps the learning on both Cup Catch and Reacher Easy tasks.

\subsection{Ablation Study for Visual Control} \label{appendix: pixel, ablation}

\begin{figure*}[t]
\centering
\includegraphics[width=0.95\textwidth, ]{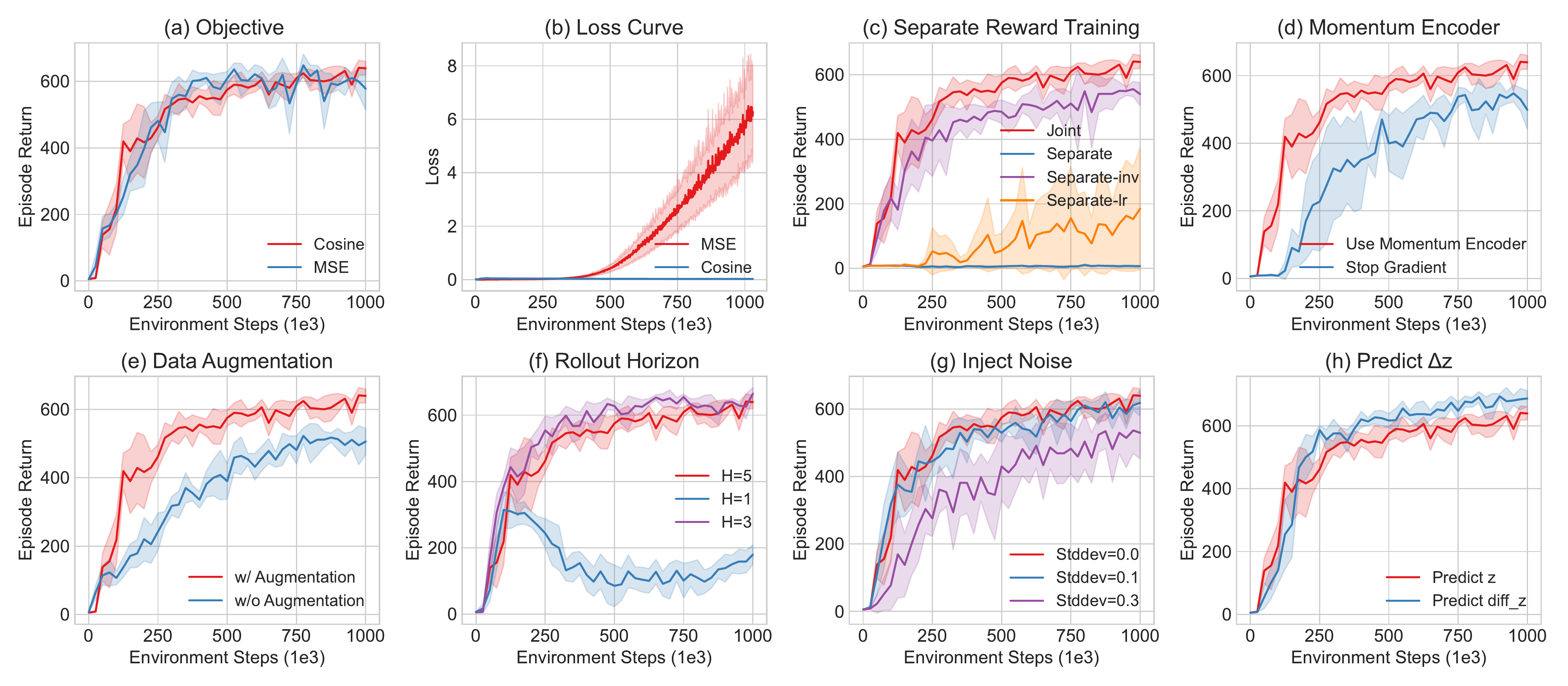}
\caption{Ablation results on pixels-based tasks. The experiments are conducted on the Cheetah Run task.}
\label{fig: ablation, pixel}
\end{figure*}

\paragraph{Representation collapse in pixel-based tasks}
In our method, perhaps the most urgent concern is whether there is a representation collapse, that is, whether there exists a trivial solution where the minimal loss is obtained when the encoder maps all observations into the same constant and the transition function is identical. In Figure \ref{fig: ablation, pixel}(a-d), with Cheetah Run tasks, we investigate three factors: i) the loss function (MSE loss vs.\ cosine loss), ii) the jointly trained reward function and iii) the momentum encoder.

Unlike \cite{DBLP:conf/iclr/SchwarzerAGHCB21}, when using the MSE loss over latent states, we do not observe representation collapse and the control performance is reasonable (See Figure \ref{fig: ablation, pixel}(a-b)). However, the loss keeps growing, making training potentially unstable as discussed in section \ref{ablation: objective}, while the cosine loss nicely behaves, which is aligned with our conclusion in state-based tasks, as in section \ref{ablation: objective}.

When the reward function is trained separately from the encoder and the transition function (Figure \ref{fig: ablation, pixel}c), representation collapse happens. \citet{tian2021understanding,grill2020bootstrap} mention that using a near-optimal projector avoids representation collapse, thus we adopt different learning rates for the transition model and the rest, specifically, the learning rate of the transition model is 5 times larger than the rest. It helps somehow, but in some runs, the collapse still happens. Furthermore, when we jointly train the inverse dynamic model with the encoder and the transition model, the representation collapse is successfully prevented. Our suggestion is to jointly train the latent dynamic model with the reward function whenever available and use the inverse dynamic model if necessary.

In Figure \ref{fig: ablation, pixel}(d), when calculating the target latent states $\tilde{z}_t$ we compare usage of the momentum encoder $\tilde{z}_t = \textbf{e}_{\theta^-}(o_t)$ 
by instead using the online encoder $\tilde{z}_t = \text{stop\_grad}(\textbf{e}_{\theta}(o_t))$. We find that when the reward function is trained jointly, the second method is still able to avoid collapse but control performance drops.

\paragraph{What matters for model learning in pixel-based tasks?}
In Figure \ref{fig: ablation, pixel}(e-h), we investigate several factors that are important to the model's performance, including (i) data augmentation, (ii) rollout length during training, (iii) injecting noise into latent states, and (iv) predicting the latent state difference $\hat{z}_{t+1}-\hat{z}_t$. 

We find that data augmentation and multi-step prediction are critical to performance in pixel-based tasks, which is aligned with previous methods~\cite{DBLP:conf/iclr/SchwarzerAGHCB21,DBLP:conf/icml/HansenSW22}. Especially when using one-step prediction error during training ($H=1$), the control performance drops dramatically, suggesting the advantage of using multi-step prediction. \citet{nguyen2021temporal} injects Gaussian noise to the latent state to smooth the dynamics, however, we do not find obvious benefits of using it. Also, injecting noise $\epsilon \sim \mathcal{N}(0, 0.3^2)$ as in \citet{nguyen2021temporal} hurts the performance. Furthermore, predicting the temporal difference of $\Delta z = z_{t+1} - z_{t}$ is slightly helpful but is not the major factor.

\section{More Ablation Study in State-based Tasks} \label{appendix: ablation}
\paragraph{Hyperparameters} \label{appendix: hyperparameters}
\begin{figure*}[t]
\centering
\includegraphics[width=0.95\textwidth, ]{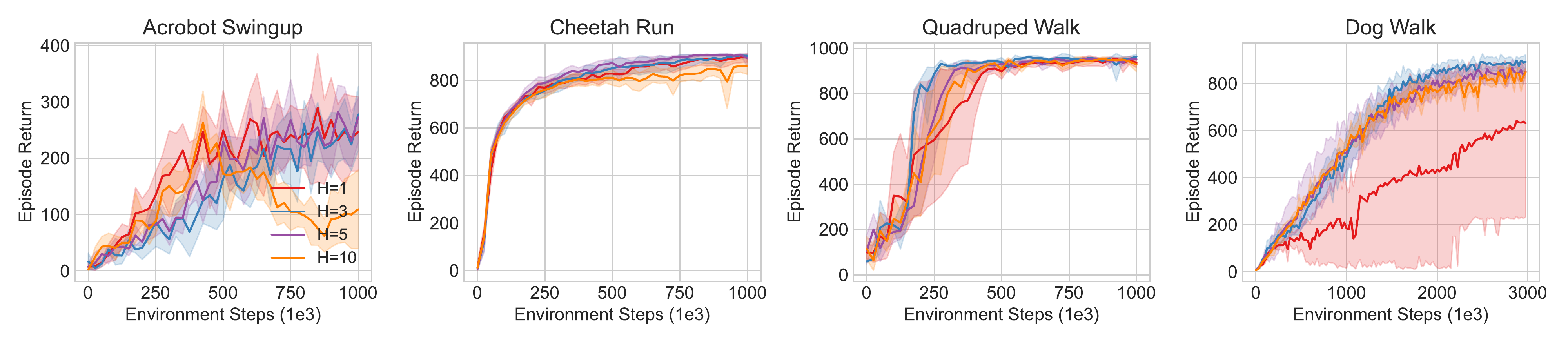}

\caption{Comparison of model rollouts horizon ($H$) during training. We vary it from $[1, 3, 5, 10]$, and we find $H=5$ works the best.}
\label{fig: ablation, horizon}
\end{figure*}

\begin{figure*}[t]
\centering
\includegraphics[width=0.95\textwidth, ]{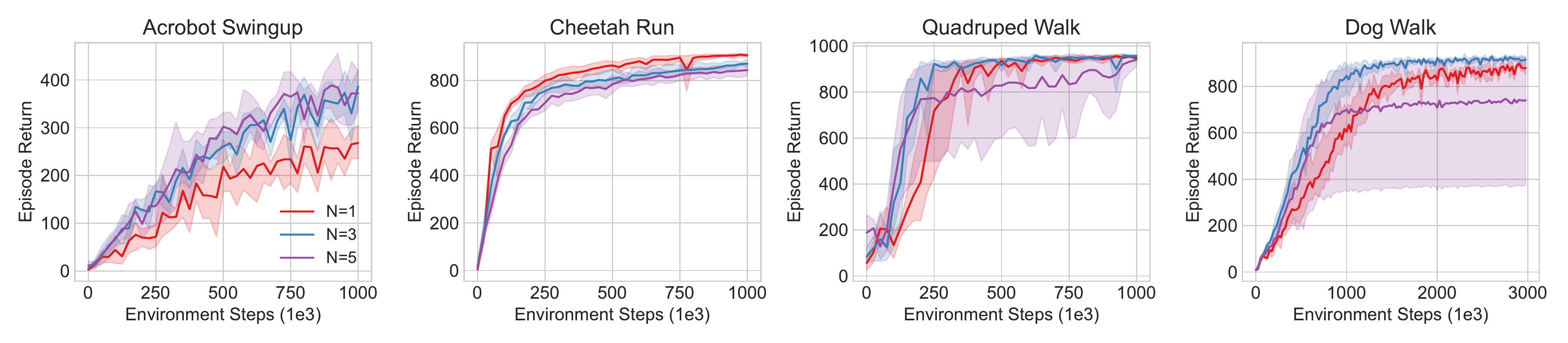}
\caption{Comparison of choosing different $N$ in N-step TD learning. We vary if from $[1, 3, 5]$ and we find $N=3$ works the best.}
\label{fig: ablation, nstep}
\end{figure*}

We incrementally show each of the influences of three hyperparameters: model rollout horizon, n-step TD learning and coefficient of the consistency loss. We first decide the model rollout horizon during training, which is noted as $H$ in Equation \ref{eq: dynamics}. We vary it from $[1, 3, 5, 10]$. As shown in Fig \ref{fig: ablation, horizon}, we find that when $H=1$, the performance decrease in the Dog Walk task, but if setting it as $H=10$, performance drops are observed in Acrobot Swingup. Thus we set it as 5 in our experiments.

We use N-step TD learning as in Equation \ref{eq: q}. Now we compare the influences of choosing different $N$ by varying it from $[1, 3, 5]$. From Figure \ref{fig: ablation, nstep}, we find that using $N=3, 5$ improves the performance in Acrobot Swingup, and using $N=5$ hurts the performance in Quadruped Walk and Dog Walk. Thus we choose $N=3$ in our experiments.

\begin{figure*}[t]
\centering
\includegraphics[width=0.95\textwidth, ]{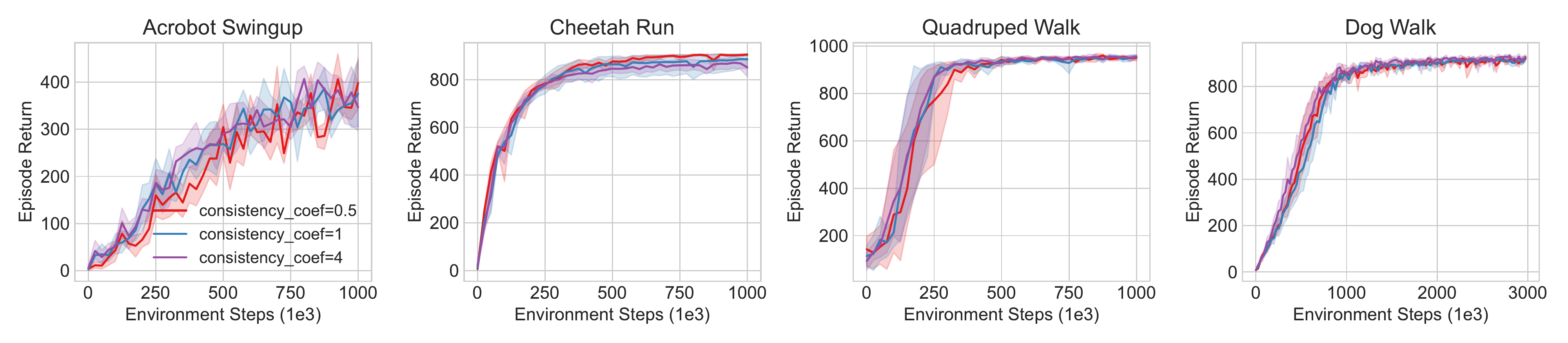}

\caption{Comparison of different coefficients of the consistency\_loss. We vary it from $[0.5, 1, 4]$ and we find the performance is robust w.r.t this hyperparameter. Thus we choose it as 1 in our experiments.}
\label{fig: ablation, coef}
\end{figure*}

Lastly, we decide on the coefficient of the consistency loss in Equation \ref{eq: dynamics}. We fix the weight of reward loss as 1 and test the coefficient of the consistency loss from $[0.5, 1, 4]$, as shown in Figure \ref{fig: ablation, coef}. We find that the algorithm is robust w.r.t this hyperparameter, so we select it as 1 in our experiment.

\paragraph{Different objective functions}
Our method exploits temporal consistency, and we compared it (TCRL-tc) with the contrastive objective in the pixel-based tasks. We now compare it with other training objectives in state-based tasks. Specifically, we compared it with i) reconstructing the observations, named TCRL-rec, and ii) without adding a loss function on the latent states, where the latent dynamic model is learned to accurately predict future rewards, named TCRL-no. As shown in Figure \ref{fig: ablation, rec, no}, our method using the temporal consistency loss works the best. Other objectives can also achieve reasonable performance on medium-level tasks, but TCRL is the only method that is able to solve the Dog Walk task (achieve more than 800 episodic returns). This experiment shows the effectiveness of leveraging temporal consistency in RL.

\begin{figure*}[t]
\centering
\includegraphics[width=0.95\textwidth, ]{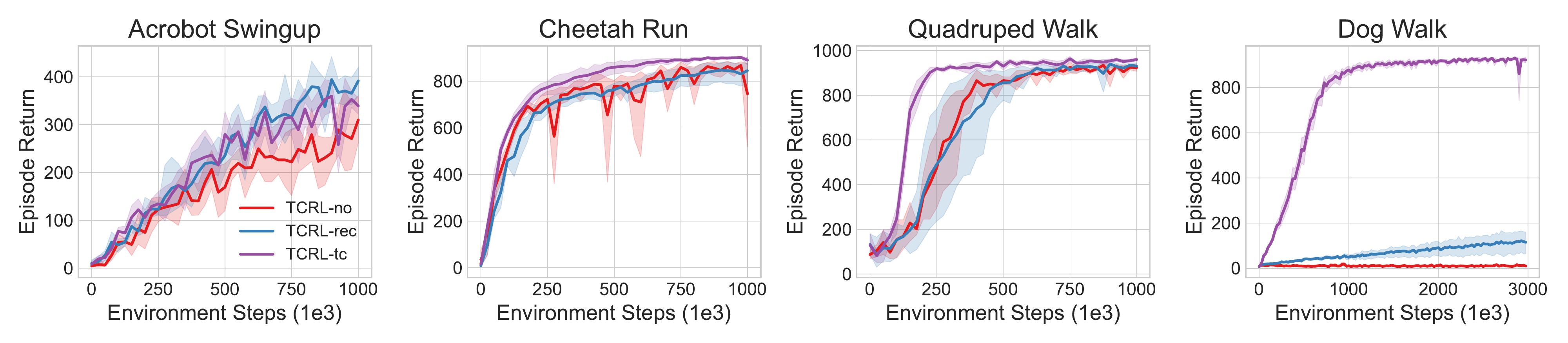}

\caption{Comparison of different objective functions. We compare our method using temporal consistency loss with reconstruction loss (TCRL-rec) and dropping the loss by predicting rewards only (TCRL-no). All methods work reasonably on medium-level tasks but only our method can solve the complex Dog Walk task.}
\label{fig: ablation, rec, no}
\end{figure*}

\section{Baselines}
\subsection{More Baselines} \label{appendix: }
We further compare our method with following baselines:
\begin{itemize}
    \item \textbf{TD3}-like method is a strong model-free baseline. It is used in DrQv2~\cite{DBLP:conf/iclr/YaratsFLP22}, with two major differences from the original TD3 method~\cite{fujimoto2018addressing}: i) it uses N-step TD learning and ii) it doesn't use the target policy network. We use it as the backend in TCRL to learn the policy and Q functions, thus the major difference between this baseline and TCRL is whether to use temporal consistency training. We include this baseline to stress the influence of representation learning.
    \item \textbf{ALM}~\cite{DBLP:journals/corr/abs-2209-08466} learns the representation, latent-space dynamics and the policy jointly. The policy is updated by recurrently backpropagating stochastic gradients~\cite{heess2015learning}. We reproduce ALM's results with author's source code. We change the testing suits from OpenAI Gym~\cite{brockman2016openai} to DeepMind Control Suite~\cite{tunyasuvunakool2020} without changing hyperparameters. We fail to achieve good performance in the tested environments but this may be due to the improper hyperparameters.
    \item \textbf{Dreamer V3}~\cite{hafner2023mastering} is the latest model-based reinforcement learning method that achieves strong performance on a diverse set of domains. The results of Dreamer V3, together with DDPG~\cite{DBLP:journals/corr/LillicrapHPHETS15} and MPO~\cite{DBLP:conf/iclr/AbdolmalekiSTMH18}, shown in Table \ref{tab:dreamerv3} are from the original Dreamer V3 paper~\cite{hafner2023mastering}. Notice that Dreamer V3 aims to work over a diverse set of domains, so the hyperparameters may not be optimized for continuous control tasks used in DMC, which potentially leads to lower performance.
    
\end{itemize}

Compared to the TD3-like baseline, we notice that TCRL hurts the performance on the Acrobot Swingup task and slightly on the Fish Swim task. However, TCRL outperforms the TD3-like baseline by a large margin on complex Humanoid and Dog domains.

\begin{figure*}[t]
\centering
\includegraphics[width=0.95\textwidth, ]{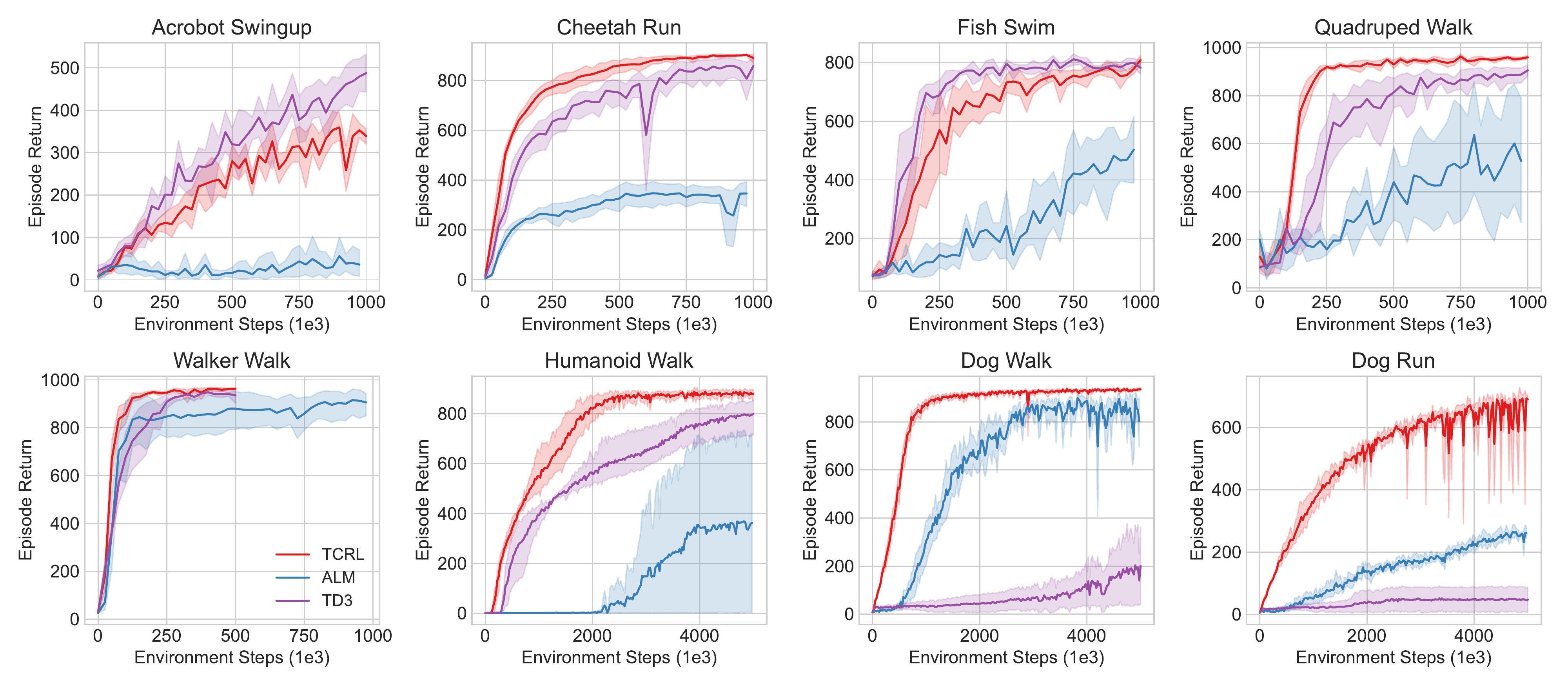}

\caption{Policy performance on eight DMC tasks compared to ALM and TD3. We plot 5 random seeds with 95\% confidence intervals presented by shaded areas.}
\end{figure*}

\begin{table*}[ht]
\centering
\caption{Policy performance on 15 DMC tasks compared with DDPG, MPO and Dreamer V3.}
\begin{tabular}{lcccc}
\hline
Task                    & DDPG     & MPO     & Dreamer V3 & TCRL \\ \hline
Acrobot Swingup         & 92.7     & 80.6    & 154.5      & \textbf{279.6} ($\sigma$ 52.1)     \\
Cartpole Swingup        & \textbf{863.9}    & \textbf{857.7}  & \textbf{850.0}      & \textbf{860.3} ($\sigma$ 4.7) \\
Cartpole Swingup Sparse & \textbf{627.5}    & 519.9   & 468.1      & \textbf{624.0} ($\sigma$ 255.0) \\ 
Cheetah Run             & 576.9	    & 612.3	 & 575.9	  & \textbf{860.7} ($\sigma$ 32.3)\\
Cup Catch               & 905.5	    & 800.6	 & \textbf{958.2}	 & \textbf{976.3} ($\sigma$ 2.1)\\
Finger Spin             & 753.6	    & 766.9	 & \textbf{937.2}	 & 849.1 ($\sigma$ 13.3) \\
Finger Turn Easy        & 462.2	    & 430.4	 & \textbf{745.4}	 & 597.9($\sigma$ 228.1)\\
Finger Turn Hard        & 286.3	    & 250.8	 & \textbf{841.0}	 & 487.8 ($\sigma$ 339.9)\\
Hopper Hop              & 24.6	    & 37.5	 & 111.0	 & \textbf{146.2} ($\sigma$ 61.4)\\
Hopper Stand            & 388.1	    & 279.3	 & 573.2	 & \textbf{664.8} ($\sigma$ 313.7)\\
Pendulum Swingup        & 748.3	    & \textbf{829.8}	 & 766.0	& \textbf{830.3} ($\sigma$ 21.0) \\
Reacher Easy            & \textbf{921.8} & \textbf{954.4}	 & \textbf{947.1}	& \textbf{938.5} ($\sigma$ 88.3 )\\
Reacher Hard            & \textbf{944.2}	    &\textbf{914.1}	 & \textbf{936.2}	& \textbf{935.7} ($\sigma$ 66.1)\\
Walker Run              & 530.0	   & 539.5	 & 632.7	 & \textbf{717.7} ($\sigma$ 46.7)\\
Walker Walk             & \textbf{948.7} & \textbf{924.9}	 & \textbf{935.7}	& \textbf{955.6} ($\sigma$ 19.0)\\ \hline
Mean                    & 605.0	    & 586.6	 & 695.5	& \textbf{715.0}\\
Medium                  & 627.5	    & 612.3	 & 766.0	& \textbf{830.3}\\
\hline
\end{tabular}
\label{tab:dreamerv3}
\end{table*}

\subsection{Extended Description of Baselines} \label{appendix: baseline}
In this section, we describe how to obtain the baseline results and what modifications are made to have fair comparisons among different algorithms in our experiments.

\begin{itemize}
    \item PETS: We implement PETS by referencing the public codebases\footnote{Code, Library of model-based RL: \href{https://github.com/facebookresearch/mbrl-lib}{https://github.com/facebookresearch/mbrl-lib}}~\footnote{Code, MBPO : \href{https://github.com/zhaoyi11/mbpo-pytorch}{https://github.com/zhaoyi11/mbpo-pytorch}}. We align hyperparemeters with our TCRL-dynamics, but used a ensemble of dynamics models. To have similar amount of parameters of each model as our method, the reward and transition function share the common two-layer MLPs $[512, 512]$ and use separate heads with two-layer MLPs $[512, 512]$. 
    \item TD-MPC: We test TD-MPC using their original code\footnote{Code, TD-MPC: \href{https://github.com/nicklashansen/tdmpc}{https://github.com/nicklashansen/tdmpc}} by changing the learning rate from 1e-3 to 3e-4 and set the update frequency from one update per environment step to every two environment steps. We also enlarge the encoder from [256, 256] to [512, 512] to have same architecture as TCRL.
    \item SAC: We obtain SAC's results by running the code implemented with Pytorch\footnote{Code, SAC: \href{https://github.com/denisyarats/pytorch_sac}{https://github.com/denisyarats/pytorch\_sac}} and make a few changes to have a fair comparison. We change the architecture of the actor and critic networks from [1024, 1024] to [512, 512, 512], add LayerNorm and Tanh nonlinear functions after the first layer according to the recommendations from \citet{furuta2021co}. We further replace the ReLU nonlinear function with ELU and change batch size from 1024 to 512. Furthermore, we set action repeat as two. 
    \item REDQ: We implement REDQ by modifying the SAC's implementation with the reference of author's implementation\footnote{Code, REDQ: \href{https://github.com/watchernyu/REDQ}{https://github.com/watchernyu/REDQ}}. We set the update-to-data ratio as 10 and reduce it to 1 for the Fish Swim task since performance collapse is observed on this task with a high ratio (10).
    \item ALM: We obtain the ALM's results by re-running the authors' implementation\footnote{Code, ALM: \href{https://github.com/RajGhugare19/alm/tree/7f1afdfd92f212a9deaf81e47e8b529b4aec2ee0}{https://github.com/RajGhugare19/alm/tree/7f1afdfd92f212a9deaf81e47e8b529b4aec2ee0}}. Except changing the testing environments from OpenAI Gym to DMC, we change the update frequency from one update per environment step to every two environment steps. We also increase the latent dimension of Humanoid and Dog to 100.
\end{itemize}

\section{Hyperparameters}
In this section, we list important hyparparameters used in both TCRL and TCRL-dynamics. For details, please check the released code\footnote{Code, TCRL: \href{https://github.com/zhaoyi11/tcrl}{https://github.com/zhaoyi11/tcrl}}. For TCRL-dynamics, we use the same hyperparameters as TCRL for learning the encoder and the latent dynamics model, thus we only list additional hyperparameters used for planning.

\begin{table}[h]
\centering
\caption{Important Hyperparameters used in TCRL and TCRL-dynamics.}
\begin{tabular}{lllll}
\cline{1-2}
\textbf{Hyperparameter}                    & \textbf{Value}  \\ \cline{1-2} 
\textbf{TCRL} & \\
Seed episode & 10 \\
Action repeat & 2 \\
Update frequency & 2 \\ 
Replay buffer size & Unlimited \\
Replay sampling strategy & Uniform \\
Batch size                        & 512    \\
Learning rate & 3e-4 \\
Optimizer & Adam \\
MLPs & [512, 512] \\
MLP activation & ELU \\
Latent Dimension & 100 (Dog, Humanoid) \\
                  &  50 (otherwise) \\
Momentum coefficient ($\tau$)            & 0.005  \\ 
Discount ($\gamma$)               & 0.99   \\
Rollout horizon ($H$)                          & 5      \\
Rollout discount & 0.9      \\
N-step TD                            & 3      \\
Reward coefficient & 1 \\
Temporal coefficient & 1 \\
Policy stddev schedule & Linear(1.0, 0.1, 50) (easy) \\
                    & Linear(1.0, 0.1, 150) (medium) \\
                    & Linear(1.0, 0.2, 500) (hard) \\
Policy stddev clip  & 0.3 \\
\cline{1-2}
\textbf{TCRL-dynamics} & \\
Planner & MPPI \\
Plan horizon ($H$) & 12 \\
Population size & 512 \\ 
Num. of elite & 64 \\
Iteration & 6 \\
Temperature & 0.5 \\
Momentum & 0.1 \\
Reuse solution & True \\
Action repeat & 2 (Dog, Walker, Finger)\\
              & 4 (Reacher, Quadruped, Cheetah) \\
              & 6 (Cup) \\
              & 8 (Cartpole) \\

\cline{1-2}
\end{tabular}

\label{tab:hyperparameter}
\end{table}

\section{Full Results}
\label{result: full}
\begin{figure}[h]
\centering
\includegraphics[width=0.95\textwidth, ]{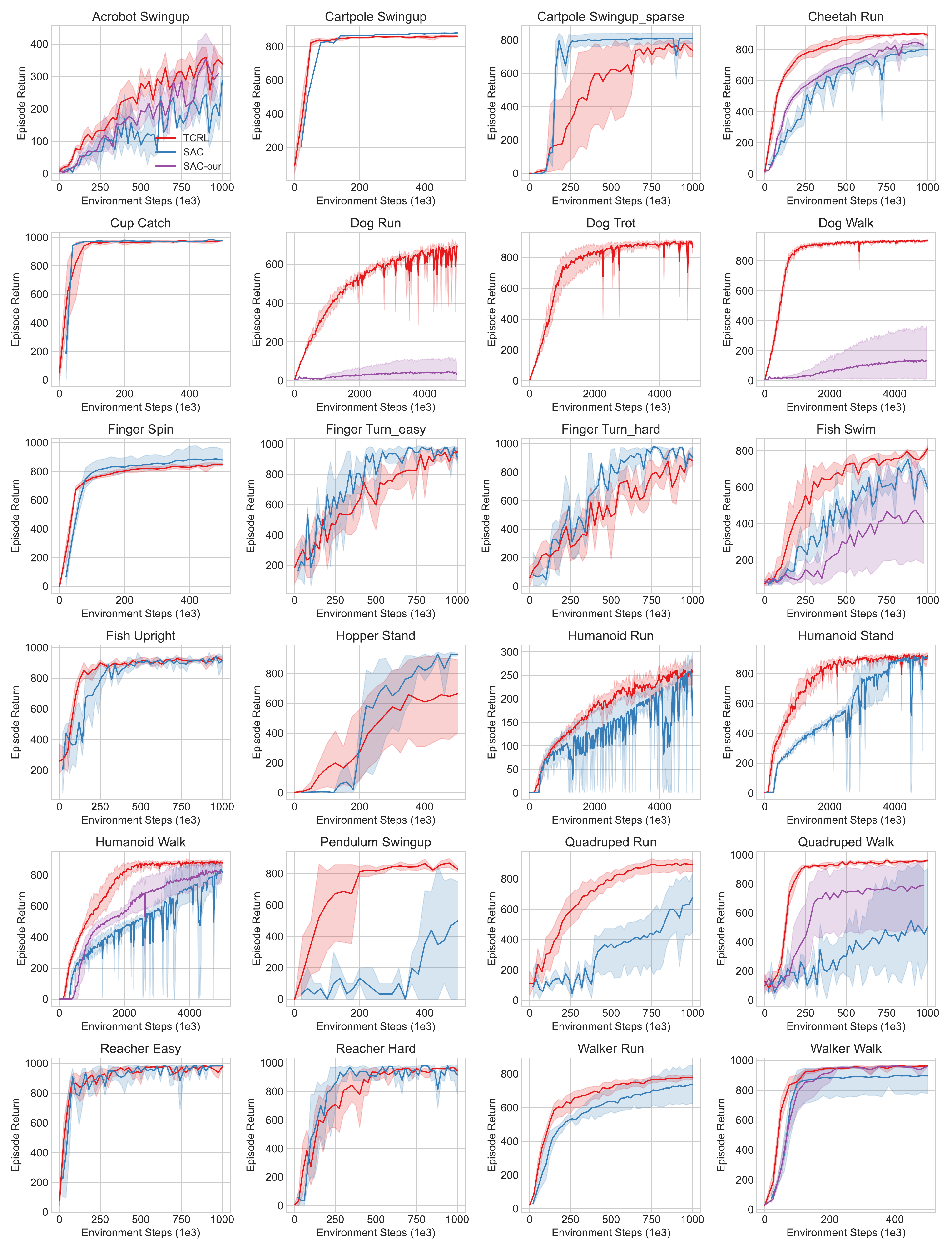}

\caption{TCRL's results on 24 continuous control tasks from DMC. We plot 5 random seeds with 95\% confidence intervals presented by shaded areas. SAC's results are from the public GitHub repository\footnote[4]{Code, SAC: \href{https://github.com/denisyarats/pytorch_sac}{https://github.com/denisyarats/pytorch\_sac}} and SAC-our's results are from Figure~\ref{fig: result_policy}.}
\label{fig:arc}
\end{figure}





\end{document}